\documentclass[letterpaper,english]{article}
\usepackage[T1]{fontenc}
\usepackage[latin9]{inputenc}
\usepackage[active]{srcltx}
\usepackage{color}
\usepackage{babel}
\usepackage{array}
\usepackage{booktabs}
\usepackage{bbding}
\usepackage{amsmath}
\usepackage{amsthm}
\usepackage{amssymb}
\usepackage{stmaryrd}
\usepackage{graphicx}
\usepackage[unicode=true] {hyperref}

\makeatletter

\providecommand{\tabularnewline}{\\}

\usepackage{microtype}
\usepackage{graphicx}
\usepackage{subfigure}
\usepackage{booktabs} 

\usepackage[accepted]{icml2019}

\icmltitlerunning{Distributed, Egocentric Representations of Graphs for Detecting Critical Structures}

\makeatother

\begin{document}
\twocolumn[
\icmltitle{Distributed, Egocentric Representations of Graphs for\\ Detecting Critical Structures}

\begin{icmlauthorlist} \icmlauthor{Ruo-Chun Tzeng}{msft} \icmlauthor{Shan-Hung Wu}{nthu} \end{icmlauthorlist}
\icmlaffiliation{msft}{Microsoft Inc.} \icmlaffiliation{nthu}{CS Department, National Tsing Hua University, Taiwan}
 \icmlcorrespondingauthor{Shan-Hung Wu}{shwu@cs.nthu.edu.tw}

\icmlkeywords{Graph Embeddings, Convolutional Neural Networks} 
\vskip 0.3in]

\printAffiliationsAndNotice{}
\begin{abstract}
We study the problem of detecting critical structures using a graph
embedding model. Existing graph embedding models lack the ability
to precisely detect critical structures that are specific to a task
at the global scale. In this paper, we propose a novel graph embedding
model, called the Ego-CNNs, that employs the ego-convolutions convolutions
at each layer and stacks up layers using an ego-centric way to detects
precise critical structures efficiently. An Ego-CNN can be jointly
trained with a task model and help explain/discover knowledge for
the task. We conduct extensive experiments and the results show that
Ego-CNNs (1) can lead to comparable task performance as the state-of-the-art
graph embedding models, (2) works nicely with CNN visualization techniques
to illustrate the detected structures, and (3) is efficient and can
incorporate with scale-free priors, which commonly occurs in social
network datasets, to further improve the training efficiency.
\end{abstract}

\section{Introduction}

\label{sec:introduction}

A graph embedding algorithm converts graphs from structural representation
to fixed-dimensional vectors. It is typically trained in a unsupervised
manner for general learning tasks but recently, deep learning approaches
\cite{bruna2013spectral,kipf2016semi,atwood2016diffusion,duvenaud2015convolutional,li2016gated,pham2017column,gilmer2017neural,niepert2016learning}
are trained in a supervised manner and show superior results against
unsupervised approaches on many tasks such as node classification
and graph classification. 

While these algorithms lead to good performance on tasks, what valuable
information can be jointly learned from the graph embedding is less
discussed. In this paper, we aim to develop a graph embedding model
that jointly discovers the critical structures, i.e., partial graphs
that are dominant to a prediction in the task (e.g., graph classification)
where the embedding is applied to. This helps people running the task
understand the reason behind the task predictions, and is particularly
useful in certain domains such as the bioinformatics, cheminformatics,
and social network analysis, where valuable knowledge may be discovered
by investigating the found critical structures. 
\begin{figure}
\begin{centering}
\begin{tabular}{cc}
\includegraphics[width=0.2\textwidth]{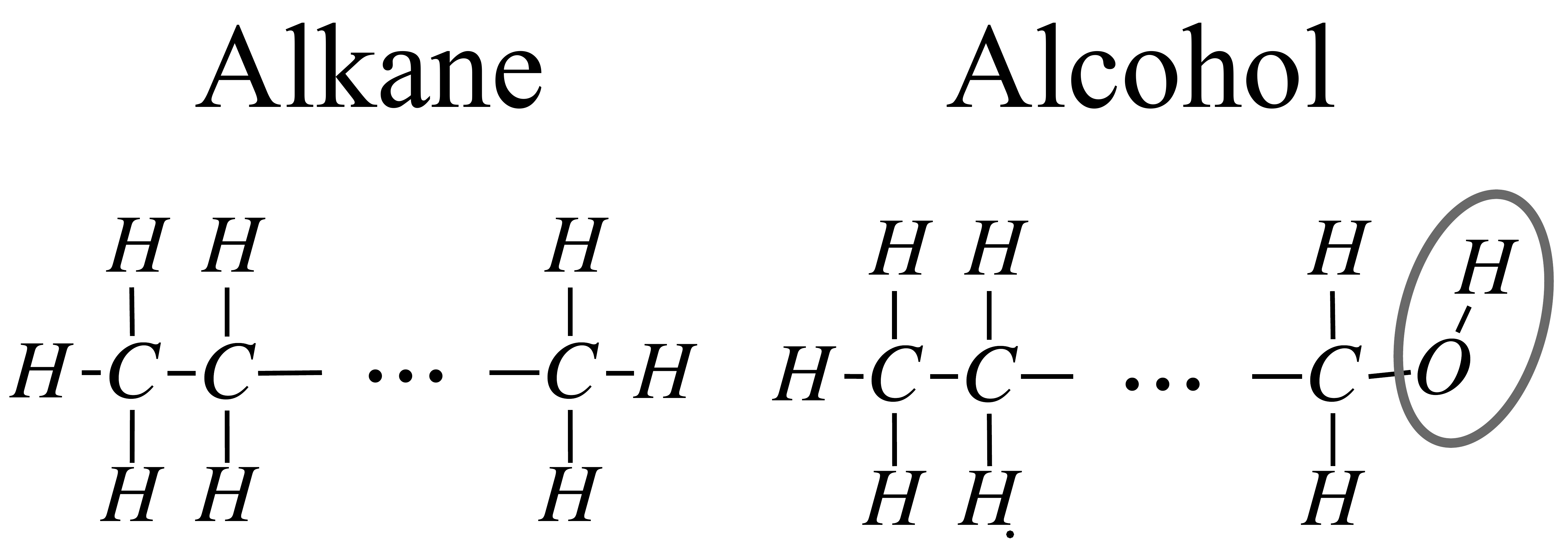} & \includegraphics[width=0.22\textwidth]{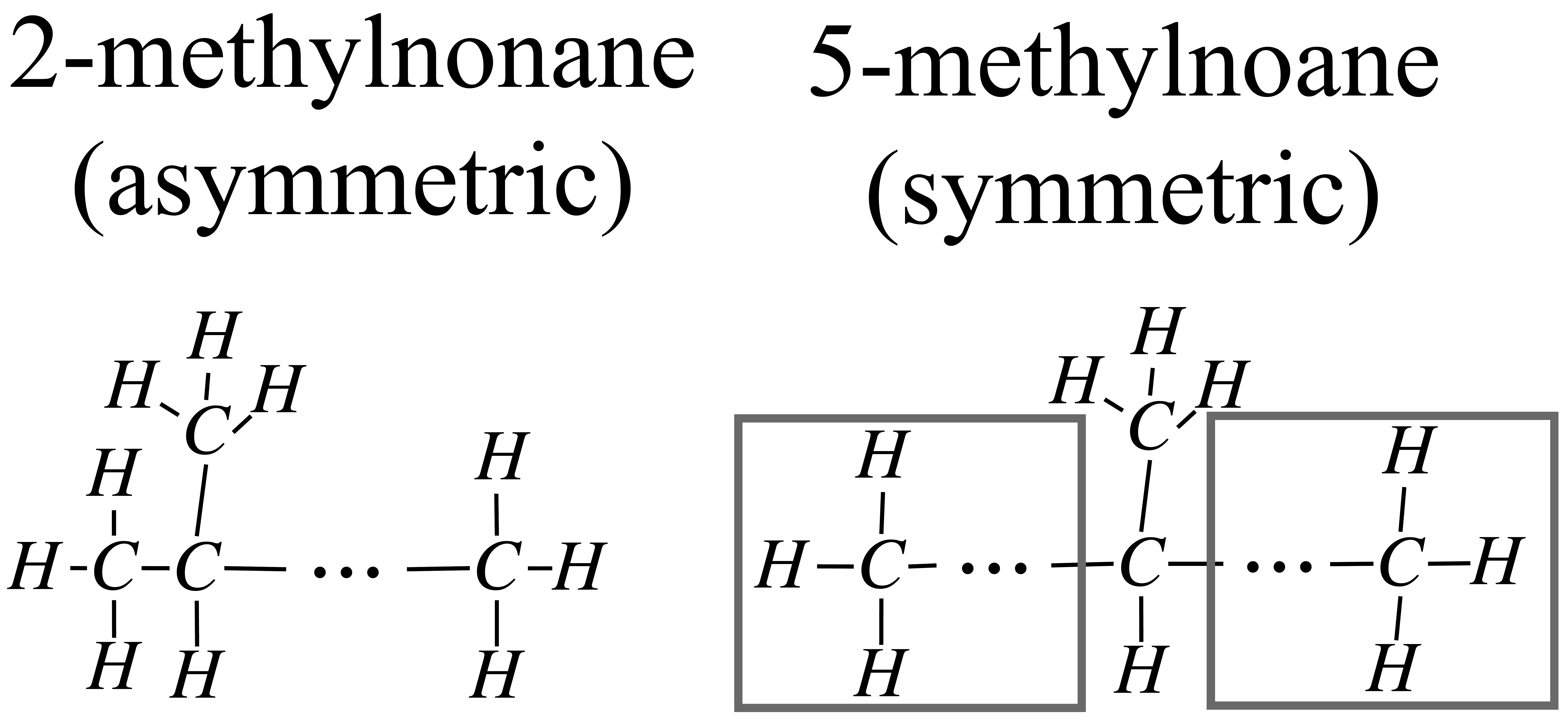}\tabularnewline
(a) & (b)\tabularnewline
\end{tabular}
\par\end{centering}
\caption{\label{fig:critical-struct-challenges}(a) The OH function group is
the critical structure to tell Alcohols from Alkanes. (b) The symmetry
hydrocarbon group (at two sides of the methyl branch) is the critical
structure to discriminate between symmetric and asymmetric isomer
of methylnonane.}
\end{figure}

However, identifying critical structures is a challenging task. The
first challenge is that critical structures are task-specific\textemdash the
shape and location of critical structures may vary from task to task.
This means that the graph embedding model should be learned together
with the task model (e.g., a classifier or regressor). The second
challenge is that model needs to be able to detect precise critical
structures. For example, to discriminant Alcohols from Alkanes (Figure
\ref{fig:critical-struct-challenges}(a)), one should check if there
exists an OH-base and if the OH-base is at the end of the compound.
To be helpful, a model has to identify the exact OH-base rather than
its approximation in any form. Third, the critical structures need
to be found at the global-scale. For example, in the task aiming to
identify if a methyl-nonane is symmetric or not (Figure \ref{fig:critical-struct-challenges}(b)),
one must check the entire graph to know if the methyl is branched
at the center position of the long carbon chain. In this task, the
critical structure is the symmetric hydrocarbon at the two sides of
the methyl branch, which can only be found at the global-scale. Unfortunately,
finding out all matches of substructures in a graph is known as subgraph
isomorphism and proven to be an NP-complete problem \cite{cook1971complexity}.
To the best of our knowledge, there is no existing graph embedding
algorithm that can identify task-dependent, precise critical structures
up to the global-scale in an efficient manner. 

\begin{table*}[t]
\caption{\label{tab:graph-embedding-model-comparison}A comparison of embedding
models for a graph $\mathcal{G}=(\mathcal{V},\mathcal{E})$, $\vert\mathcal{V}\vert=N$,
where $D$ is the embedding dimension, $K$ is the maximum node degree
in $\mathcal{G}$, $L$ is the number of layers of a deep model, and
$C$ is a graph-specific constant.}

\vspace{2.6mm}
\centering{}\resizebox{\textwidth}{!}{%
\begin{tabular}{>{\centering}m{5.5cm}|>{\centering}m{2.5cm}>{\centering}m{2.5cm}>{\centering}m{2.5cm}>{\centering}m{2.5cm}>{\centering}m{3.5cm}}
\multicolumn{1}{>{\centering}m{5.5cm}}{\textbf{Graph embedding model}} & \textbf{Task-Specific?} & \textbf{Precise critical structures?} & \textbf{Exponential scale efficiency?} & \textbf{Efficient on large graphs?} & \textbf{Time complexity}

\textbf{(forward pass)}\tabularnewline
\hline 
\hline 
WL kernel \cite{shervashidze2011weisfeiler} &  &  & \Checkmark{} & \Checkmark{} & $O(L(KN+C))$\tabularnewline
\hline 
DGK \cite{yanardag2015deep} &  &  &  & \Checkmark{} & $O(DCN)$\tabularnewline
\hline 
Subgraph2vec \cite{narayanan2016subgraph2vec} &  &  &  & \Checkmark{} & $O(DCN)$\tabularnewline
\hline 
MLG \cite{kondor2016multiscale} &  &  & \Checkmark{} &  & $O(LN^{5})$\tabularnewline
\hline 
Structure2vec \cite{dai2016discriminative} & \Checkmark{} &  & \Checkmark{} & \Checkmark{} & $O((KD+D^{2})LN)$\tabularnewline
\hline 
Spatial GCN \cite{bruna2013spectral} & \Checkmark{} & \Checkmark{} &  &  & $O(DLN^{2})$\tabularnewline
\hline 
Spectrum GCN \cite{bruna2013spectral,defferrard2016convolutional,kipf2016semi} & \Checkmark{} &  & \Checkmark{} & \Checkmark{} & $O(D^{2}L|\mathcal{E}|)$\tabularnewline
\hline 
DCNN \cite{atwood2016diffusion} & \Checkmark{} &  &  &  & $O(MDN^{2})$\tabularnewline
\hline 
Patchy-San \cite{niepert2016learning} & \Checkmark{} & \Checkmark{} &  & \Checkmark{} & $O(K^{2}DN)$\tabularnewline
\hline 
Message-Passing NNs \cite{duvenaud2015convolutional,li2016gated,pham2017column,gilmer2017neural,velickovic2018graph,ying2018hierarchical} & \Checkmark{} &  & \Checkmark{} & \Checkmark{} & $O(K^{2}D^{2}LN)$\tabularnewline
\hline 
\hline 
Ego-CNN & \Checkmark{} & \Checkmark{} & \Checkmark{} & \Checkmark{} & $O(KD^{2}LN)$\tabularnewline
\hline 
\end{tabular}}
\end{table*}

In this paper, we present the Ego-CNNs\footnote{The code is available at \href{https://github.com/rutzeng/EgoCNN}{https://github.com/rutzeng/EgoCNN}.}
that embed a graph into distributed (multi-layer), fixed-dimensional
tensors. An Ego-CNN is a feedforward convolutional neural network
that can be jointly learned with a supervised task model (e.g., fully-connected
layers) to help identify the task-specific critical structures. The
Ego-CNNs employ novel ego-convolutions to learn the latent representations
at each network layer. Unlike the neurons in most existing task-specific,
NN-based graph embedding models \cite{bruna2013spectral,kipf2016semi,atwood2016diffusion,duvenaud2015convolutional,li2016gated,pham2017column,gilmer2017neural}
which detect only fuzzy patterns, a neuron in an Ego-CNN can detect
precise patterns in the output of the previous layer. This allows
the precise critical structures to be backtracked following the model
weights layer-by-layer after training. Furthermore, we propose the
ego-centric design for stacking up layers, where the receptive fields
of neurons across layers center around the same nodes. Such design
avoids the locality and efficiency problems in existing precise model
\cite{niepert2016learning} and enables efficient detection of critical
structures at the global scale.  

We conduct extensive experiments and the results show that Ego-CNNs
work nicely with some common visualization techniques for CNNs, e.g.,
Transposed Deconvolution \cite{zeiler2011adaptive}, can successfully
output critical structures behind each prediction made by the jointly
trained task model, and in the meanwhile, achieving performance comparable
to the state-of-the-art graph classification models.  We also show
that Ego-CNNs can readily incorporate the scale-free prior, which
commonly exists in large (social) graphs, to further improve the training
efficiency in practice. To the best of our knowledge, the Ego-CNNs
are the first graph embedding model that can efficiently detect task-dependent,
precise critical structures at the global scale.

\section{Related Work}

\label{sec:related-work}

Next, we briefly review existing graph embedding models. Table \ref{tab:graph-embedding-model-comparison}
compares the Ego-CNNs with existing graph embedding approaches. For
an in-depth review of existing work, please refer to Section 1 of
the supplementary materials or the survey \cite{cai2018comprehensive}. 

Traditional graph kernels, including the Weisfeiler-Lehman (WL) kernel
\cite{shervashidze2011weisfeiler}, Deep Graph Kernels (DGKs) \cite{yanardag2015deep},
Subgraph2vec \cite{narayanan2016subgraph2vec}, and Multiscale Laplacian
Graph (MLG) Kernels \cite{kondor2016multiscale} are designed for
unsupervised tasks. They have difficulty of finding task-specific
critical structures. 

Some recent studies aim to learn the task-specific graph embeddings.
Structure2vec \cite{dai2016discriminative} uses approximated inference
techniques to embed a graph. Studies, including Spectrum Graph Convolutional
Network (GCN) \cite{bruna2013spectral} and its variant \cite{kipf2016semi},
Diffusion Convolutional Neural Networks (DCNNs) \cite{atwood2016diffusion},
and Message-Passing Neural Networks (MPNNs) \cite{duvenaud2015convolutional,li2016gated,pham2017column,gilmer2017neural,velickovic2018graph,ying2018hierarchical},
borrow the concepts of CNNs to embed graphs. The idea is to model
the filters/kernels that scan through different parts of the graph
(which we call the neighborhoods) to learn patterns most helpful to
the learning task. However, the above work share a drawback that they
can only identify fuzzy critical structures or critical structures
of very simple shapes due to the ways the convolutions are defined
(to be elaborated later in this secion). Recently, a node embedding
model, called the Graph Attention Networks (GATs) \cite{velickovic2018graph},
is proposed. The GAT attention can be 1- or multi-headed. The multi-head-attention
GATs aggregate hidden representations just like Message-Passing NNs
and thus cannot detect precise critical structures. On the other hand,
the 1-head-attention GATs allow backtracking precise nodes covered
by a neuron through their masked self-attentional layers. However,
being node embedding models, the 1-head-attention GATs can only detect
simplified patterns in neighborhoods. The Ego-CNNs are a generalization
of 1-head-attention GATs for graph embedding. We will compare these
two models in more details in Section \ref{sec:ego-cnn-model}. 

The only graph embedding models that allow backtracking nodes covered
by a neuron are Spatial GCN \cite{bruna2013spectral} and Patchy-San
\cite{niepert2016learning}. Unfortunately, the Spatial GCN is not
applicable to our problem since it aims to perform hierarchical clustering
of nodes. The filters learn the distance between clusters (a graph-level
information) rather than subgraph patterns. On the other hand, the
Patchy-San \cite{niepert2016learning} can detect precise critical
structures. However, it is a single-layer NN\footnote{Some people misunderstand that the Patchy-San has multiple layers
since the original paper \cite{niepert2016learning} used a four-layer
NN for experiment. In fact, the second layer is a traditional convolutional
layer and the latter two serve as the task model. } designed to detect only the local critical structures around each
node. Due to the lack of recursive definition of convolutions at deep
layers, the Patchy-San does not enjoy the exponential efficiency of
detecting large-scale critical structures using multiple layers as
in other CNN-based models. 

\begin{figure}
\begin{centering}
\begin{tabular}{ccc}
\includegraphics[height=2.2cm]{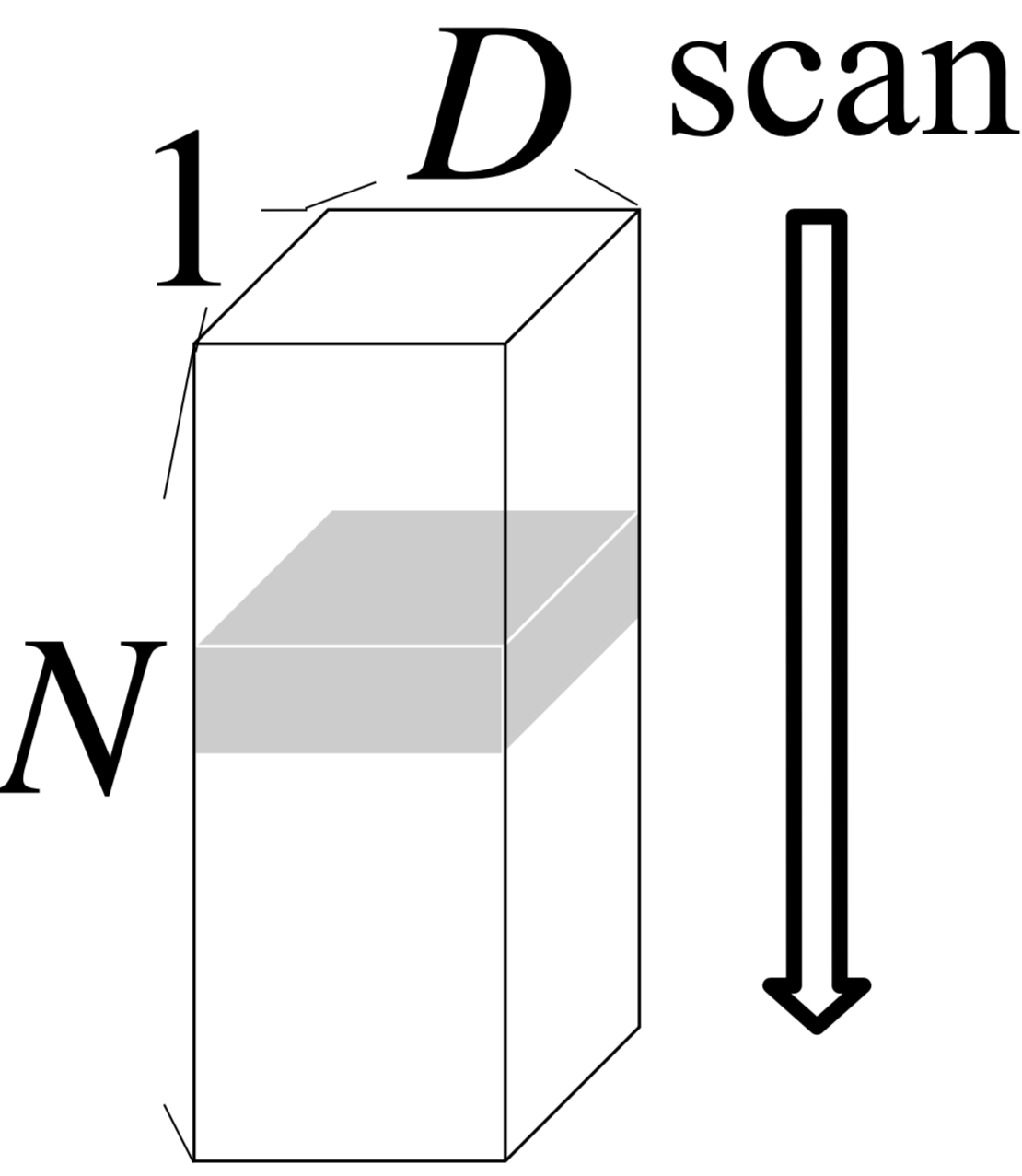} & \includegraphics[height=2.2cm]{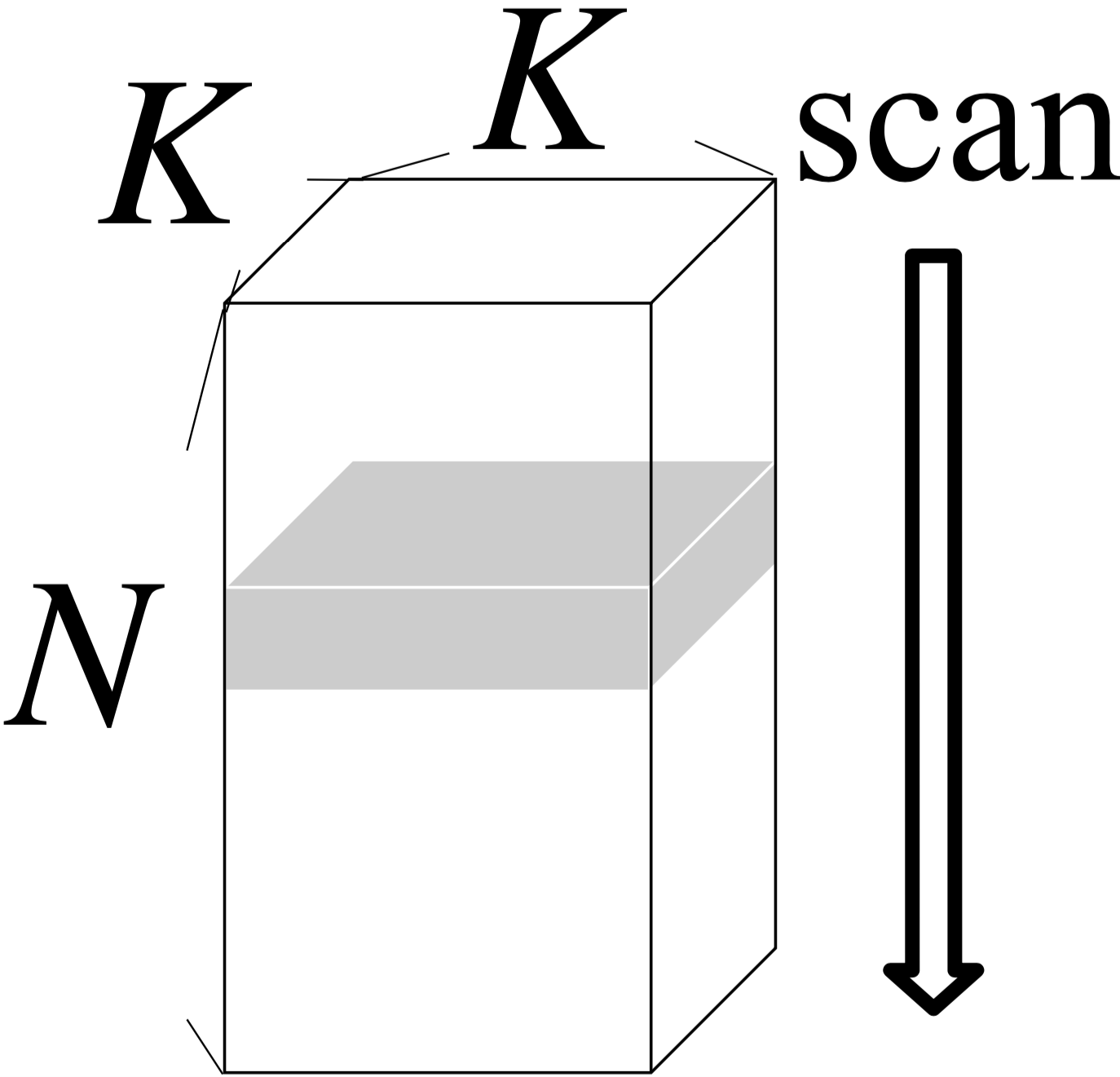} & \includegraphics[height=2.2cm]{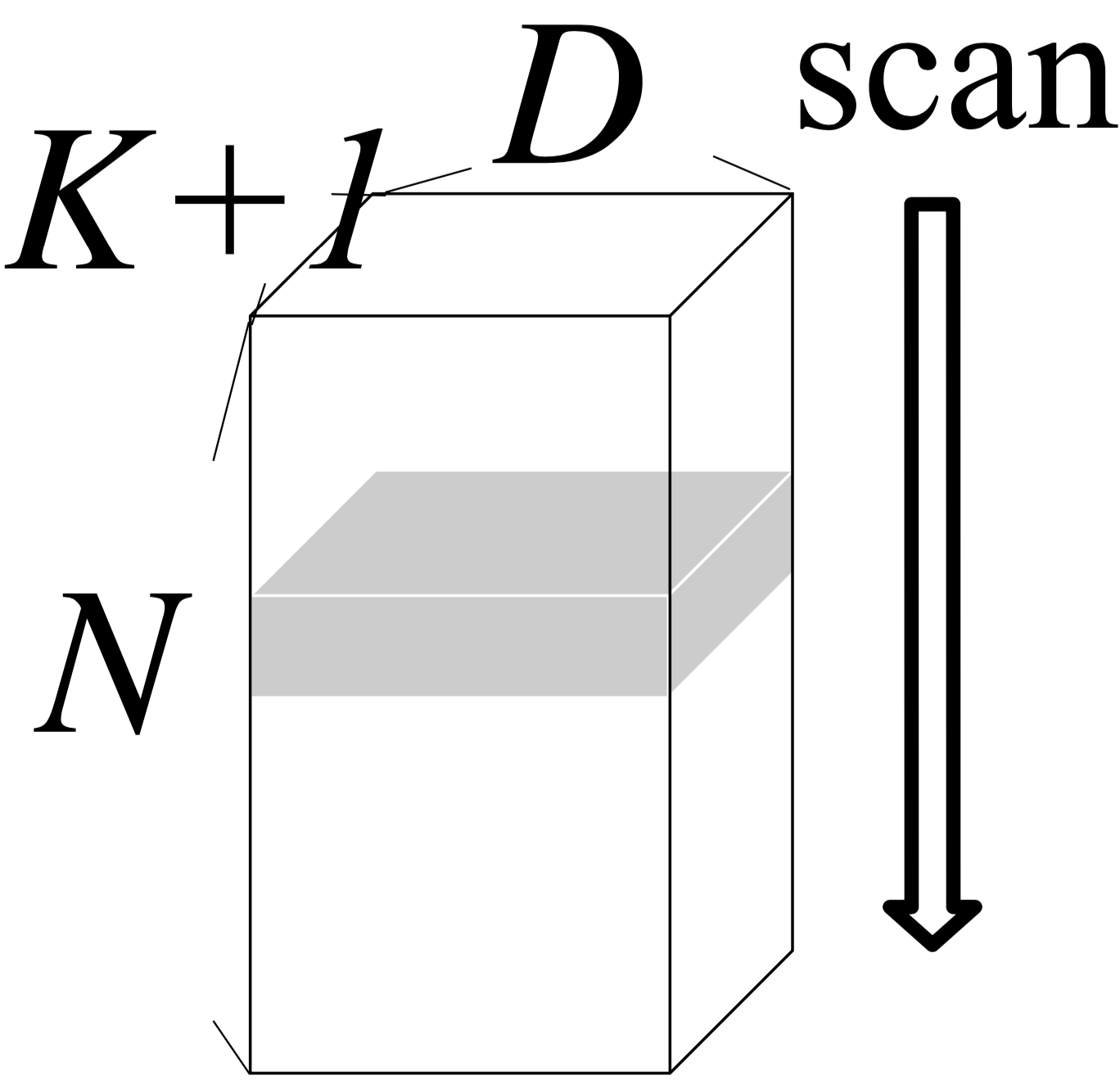}\tabularnewline
(a) & (b) & (c)\tabularnewline
\end{tabular}
\par\end{centering}
\caption{\label{fig:cnn-filters}Neighborhood of a node $n$ in (a) Message-Passing
NNs \cite{duvenaud2015convolutional,li2016gated,pham2017column,gilmer2017neural}:
$\boldsymbol{g}_{n}^{(l)}\in\mathbb{R}^{D}$ the aggregated hidden
representations of adjacent nodes in the previous layer; (b) Patchy-San
\cite{niepert2016learning}: $\boldsymbol{A}^{(n)}\in\mathbb{R}^{K\times K}$
the adjacency matrix of $K$ nearest neighbors of node $n$, and (c)
Ego-CNNs: ${\displaystyle \boldsymbol{E}^{(n,l)}}\in\mathbb{R}^{(K+1)\times D}$
the hidden representation of the $l$-hop ego network centered at
node $n$ with $K$ nearest neighbors.}
\end{figure}

\begin{figure*}
\begin{centering}
\includegraphics[width=0.9\textwidth]{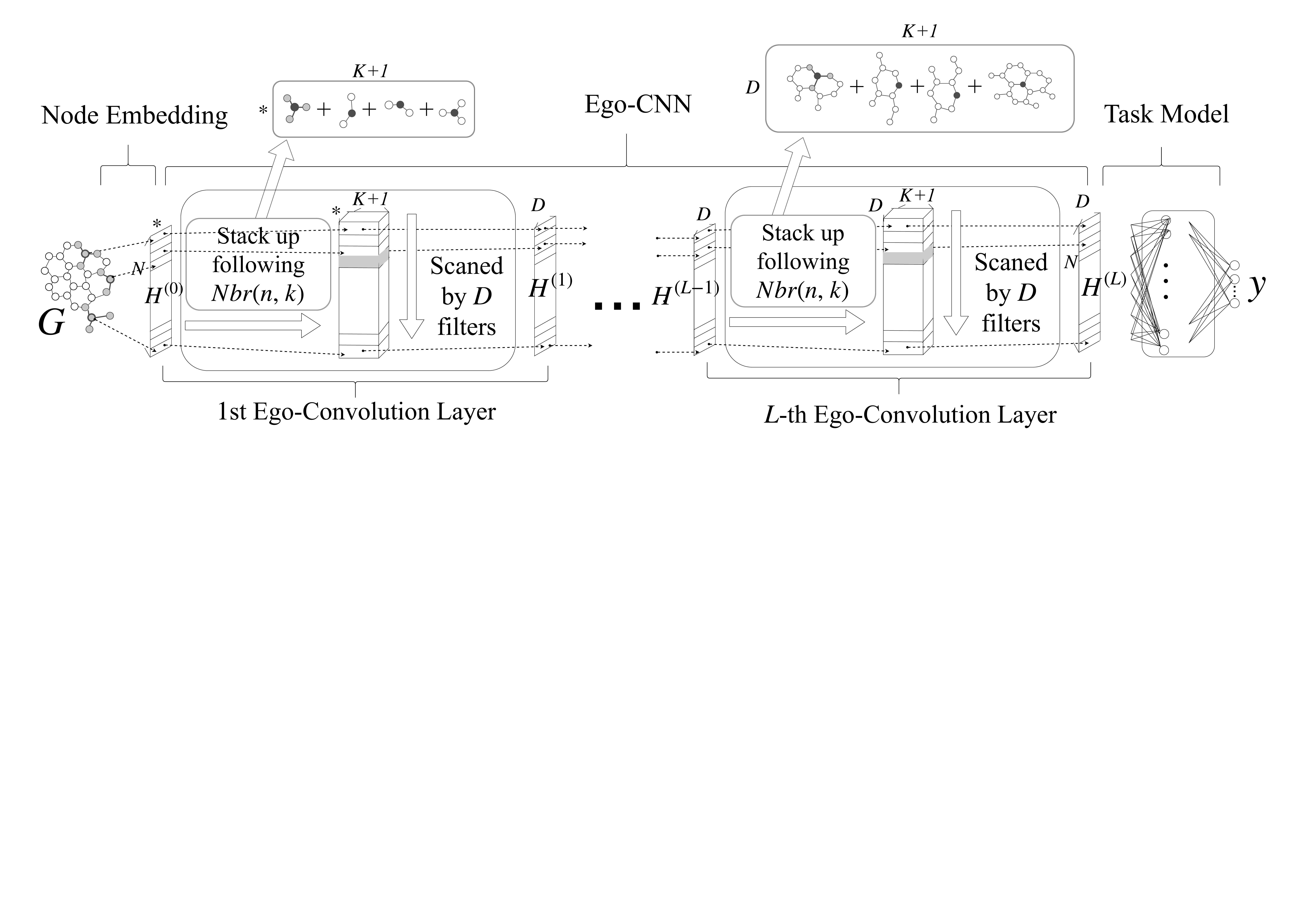}
\par\end{centering}
\caption{\label{fig:model-egocnn} The model architecture of an Ego-CNN. With
our egocentric d\textcolor{black}{esign, neighborhoods are egocentrically
enlarged by 1-hop after each Ego-Convolution layer. The dashed horizontal
lines across layers indicate neighborhoods of the same node; the ?
mark indicates an arbitrary dimension. }}
\end{figure*}

\textbf{Neighborhoods.} To understand the cause of limitations in
existing work, we look into the definitions of neighborhoods in these
models, as shown in Figure \ref{fig:cnn-filters}. In a CNN, a filters/kernel
scans through different neighborhoods in a graph to detect the repeating
patterns across neighborhoods. Hence, the definition of a neighborhood
determines what to be learned by the model. In Message-Passing NNs
\cite{duvenaud2015convolutional,li2016gated,pham2017column,gilmer2017neural}
(Figure \ref{fig:cnn-filters}(a)), the $d$-th filter $\boldsymbol{w}^{(l,d)}\in\mathbb{R}^{D}$,
$d=1,2,\cdots,D$, at the $l$-th layer scans through the $D$-dimensional
vector $\boldsymbol{g}^{(n,l)}$ of every node $n$. The vector $\boldsymbol{g}^{(n,l)}=\bigoplus_{m\in Adj(n)}\boldsymbol{h}^{(m,l-1)}\in\mathbb{R}^{D}$
is an aggregation $\bigoplus$ (e.g., summation \cite{duvenaud2015convolutional})
of the hidden representations $\boldsymbol{h}^{(m,l-1)}$'s of the
adjacent nodes $m$'s at the $(l-1)$-th layer. The $d$-th dimension
of the hidden representations $h_{d}^{(n,l)}$ of a node $n$ at the
$l$-th layer is calculated by\footnote{We have made some simplifications. For more details, please refer
to a nice summary \cite{gilmer2017neural}. }
\begin{equation}
h_{d}^{(n,l)}=\sigma(\boldsymbol{g}^{(n,l)\top}\boldsymbol{w}^{(l,d)}+b_{d}),\label{eq:h-mpnn}
\end{equation}
where $\sigma$ is an activation function and $b_{d}$ is a bias term.
By stacking up layers in these models, a deep layer can efficiently
detect patterns that cover exponentially more nodes than the patterns
found in the shallow layers. However, these models loses the ability
of detecting precise critical structures since the network weights
$\boldsymbol{w}^{(l,d)}$'s at the $l$-th layer parametrize only
the aggregated representations from the previous layer. It is hard
(if not impossible) for these networks to backtrack the critical nodes
in the $(l-1)$-th layer via the weights $\boldsymbol{w}^{(l,d)}$'s
after model training. 

The single-layer Patchy-San \cite{niepert2016learning} uses filters
to detects patterns in the adjacency matrix of the $K$ nearest neighbors
of each node. The neighborhood of a node $n$ is defined as the $K\times K$
adjacency matrix $\boldsymbol{A}^{(n)}$ of the $K$ nearest neighbors
of the node (Figure \ref{fig:cnn-filters}(b)). Filters $\boldsymbol{W}^{(d)}\in\mathbb{R}^{K\times K}$,
$d=1,2,\cdots,D$, scan through the adjacency matrix of each node
to generate the graph embedding $\boldsymbol{H}\in\mathbb{R}^{N\times D}$,
where 
\begin{equation}
{\displaystyle H_{n,d}=\sigma(\boldsymbol{A}^{(n)}\varoast\boldsymbol{W}^{(d)}+b_{d}})\label{eq:h-patchy-san}
\end{equation}
is the output of an activation function $\sigma$, $b_{d}$ is the
bias term, and $\varoast$ is the Frobenius inner product defined
as $\boldsymbol{X}\varoast\boldsymbol{Y}={\displaystyle \varSigma_{i,j}X_{i,j}Y_{i,j}}$.
 Unlike in Message-Passing NNs, the filters $\boldsymbol{W}^{(d)}$'s
in Patchy-San parametrize the non-aggregated representations of nodes.
Thus, by backtracking the nodes via $\boldsymbol{W}^{(d)}$'s, one
can discover precise critical structures. However, to detect critical
structures at the global scale, each $\boldsymbol{A}^{(n)}$ needs
to have the size of $N\times N$, making the filters $\boldsymbol{W}^{(d)}\in\mathbb{R}^{N\times N}$
hard to learn. Efficient detection of task-specific, precise critical
structures at global scale remains an important but unsolved problem. 

\section{Ego-CNN}

\label{sec:ego-cnn}A deep CNN model, when applied to images, offers
two advantages: (1) filers/kernels at a layer, by scanning the neighborhood
of every pixel, detect location independent patterns, and (2) with
a proper recursive definition of neighborhoods, a filter at a deep
layer can reuse the output of neurons at the previous layer to efficiently
detect patterns in pixel areas, called receptive fields, that are
exponentially larger (in number of pixels) than those in a shallow
layer, thereby overcoming the curse of dimensionality. We aim to keep
these advantages on graphs when designing a CNN-based graph embedding
model. 

Since Patchy-San \cite{niepert2016learning} can detect precise critical
structures at the local scale, it seems plausible to extend the notion
of its neighborhoods to deep layers. In Patchy-San, the neighborhood
of a node $n$ at the input (shallowest) layer is defined as the $K\times K$
adjacency matrix $\boldsymbol{A}^{(n)}$ of the $K$ nearest neighbors
of the node. We can recursively define the neighborhood of the node
$n$ at a deep layer $l$ as the $K\times K$ adjacency matrix $\boldsymbol{A}^{(n,l)}$
of the $K$ nearest neighbors having the most similar latent representations
output from the previous layer $(l-1$). 

However, this naive extension suffers from two drawbacks. First, the
neighborhood is dynamic since the $K$ nearest neighbors may change
during the training time. This prevents the filters from learning
the location independent patterns. Second, as the neighborhoods at
layer $(l-1)$ are dynamic, it is hard for a model designer to decide
which neuron output at layer $(l-1)$ to wire up to a filter at layer
$l$ such that the filter can reuse the output to exponentially increase
the learning efficiency in detecting large-scale patterns. As we can
see, the root cause of the above problems is the ill-defined neighborhoods.
This motivate us to rethink the definition of neighborhoods from scratch.

\subsection{Model Design}

\label{sec:ego-cnn-model}We propose the Ego-CNN model that (1) defines
ego-convolutions at each layer where a filter at layer $l$ scans
the neighborhood representing the $l$-hop ego network\footnote{In a graph, an $l$-hop ego network centered at node $n$ is a subgraph
consisting of the node and all its $l$-hop neighbors as well as the
edges between these nodes.} centered at every node, and (2) stacks up layers using an ego-centric
way such that the neighborhoods of a node $n$ at layers $1,2,\cdots,L$
center around the same node, as shown in Figure \ref{fig:model-egocnn}. 

\textbf{Ego-Convolutions.} Let $Nbr(n,k)$ be the $k$-th nearest
neighbor of a node $n$ in the graph $\mathcal{G}$ and $\boldsymbol{H}^{(l)}\in\mathbb{R}^{N\times D}$
be the graph embedding output by $D$ filters $\boldsymbol{W}^{(l,1)},\cdots,\boldsymbol{W}^{(l,D)}$
at the $l$-th layer. For $l=1,\cdots,L$, we define

\begin{equation}
\begin{array}{c}
{\displaystyle H_{n,d}^{(l)}=\sigma\left(\boldsymbol{E}^{(n,l)}\varoast\boldsymbol{W}^{(l,d)}+b_{d}^{(l)}\right)\text{, where }}\\
{\displaystyle \boldsymbol{E}^{(n,l)}=\left[\boldsymbol{H}_{n,:}^{(l-1)},\boldsymbol{H}_{Nbr(n,1),:}^{(l-1)},\cdots,\boldsymbol{H}_{Nbr(n,K),:}^{(l-1)}\right]^{\top}}
\end{array}\label{eq:model-neighborhood}
\end{equation}
The $\boldsymbol{E}^{(n,l)}\in\mathbb{R}^{(K+1)\times D}$ is a matrix
representing the neighborhood of the node $n$ at the $l$-th layer,
$\sigma$ is the activation function, $b_{d}$ is the bias term, and
$\varoast$ is the Frobenius inner product defined as $\boldsymbol{X}\varoast\boldsymbol{Y}={\displaystyle \varSigma_{i,j}X_{i,j}Y_{i,j}}$.
We determine the $K$ nearest neighbors of a node $n$ using the edge
weights (if available) or hop count\footnote{In case that two neighbors rank the same, we can use a predefined
global node ranking or the graph normalization technique \cite{niepert2016learning}
to decide the winner.} (otherwise), and define $\boldsymbol{H}_{n,:}^{(0)}\text{\ensuremath{\in\mathbb{R}}}^{K}$
as the adjacency vector between $n$ and its $K$ nearest neighbors.
The goal of the model is to learn the filters (Figure \ref{fig:cnn-filters}(c))
and bias terms at all layers that minimize the loss defined by a task. 

\begin{figure}
\begin{centering}
\par\end{centering}
\begin{centering}
\resizebox{.49\textwidth}{!}{%
\begin{tabular}{cccc}
\includegraphics[width=0.175\textwidth]{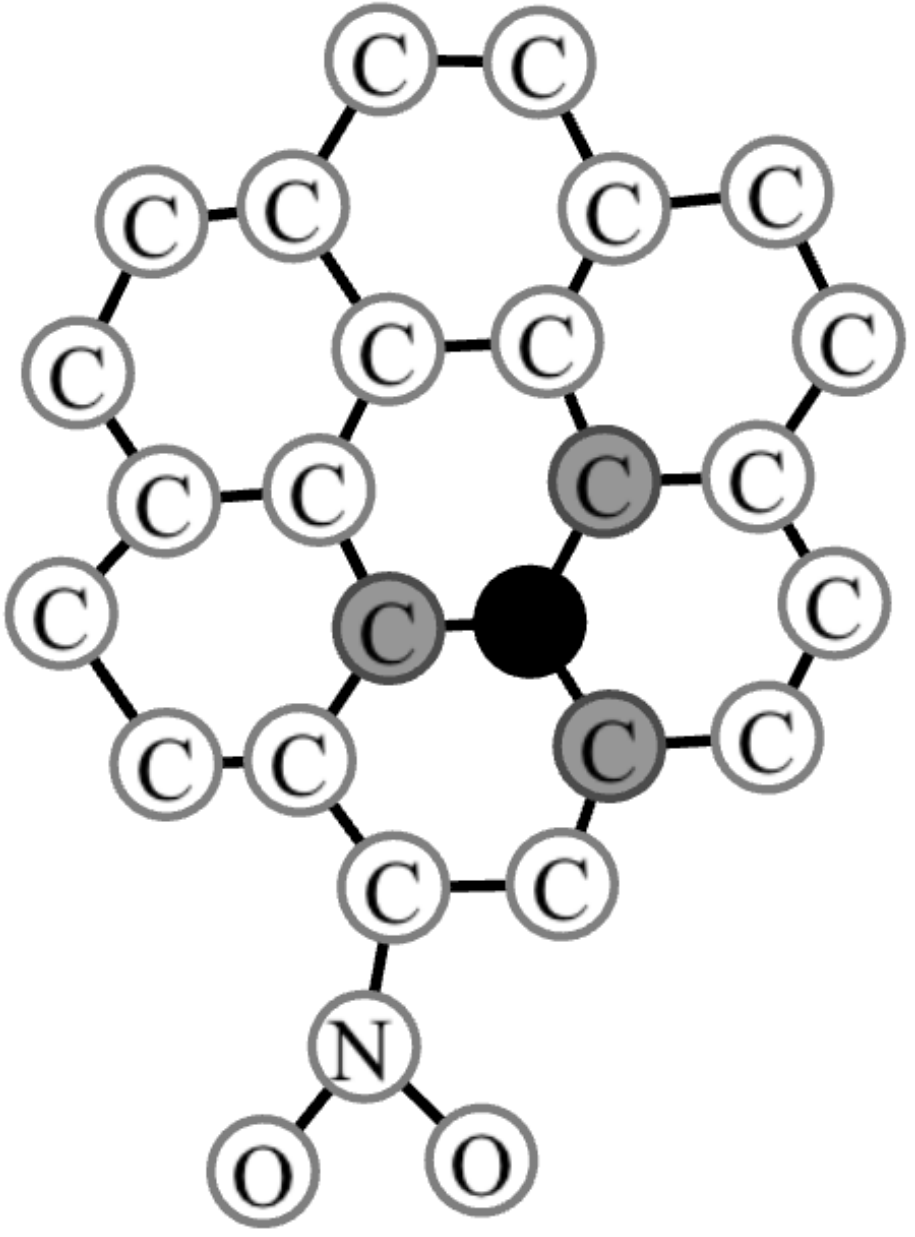} & \includegraphics[width=0.175\textwidth]{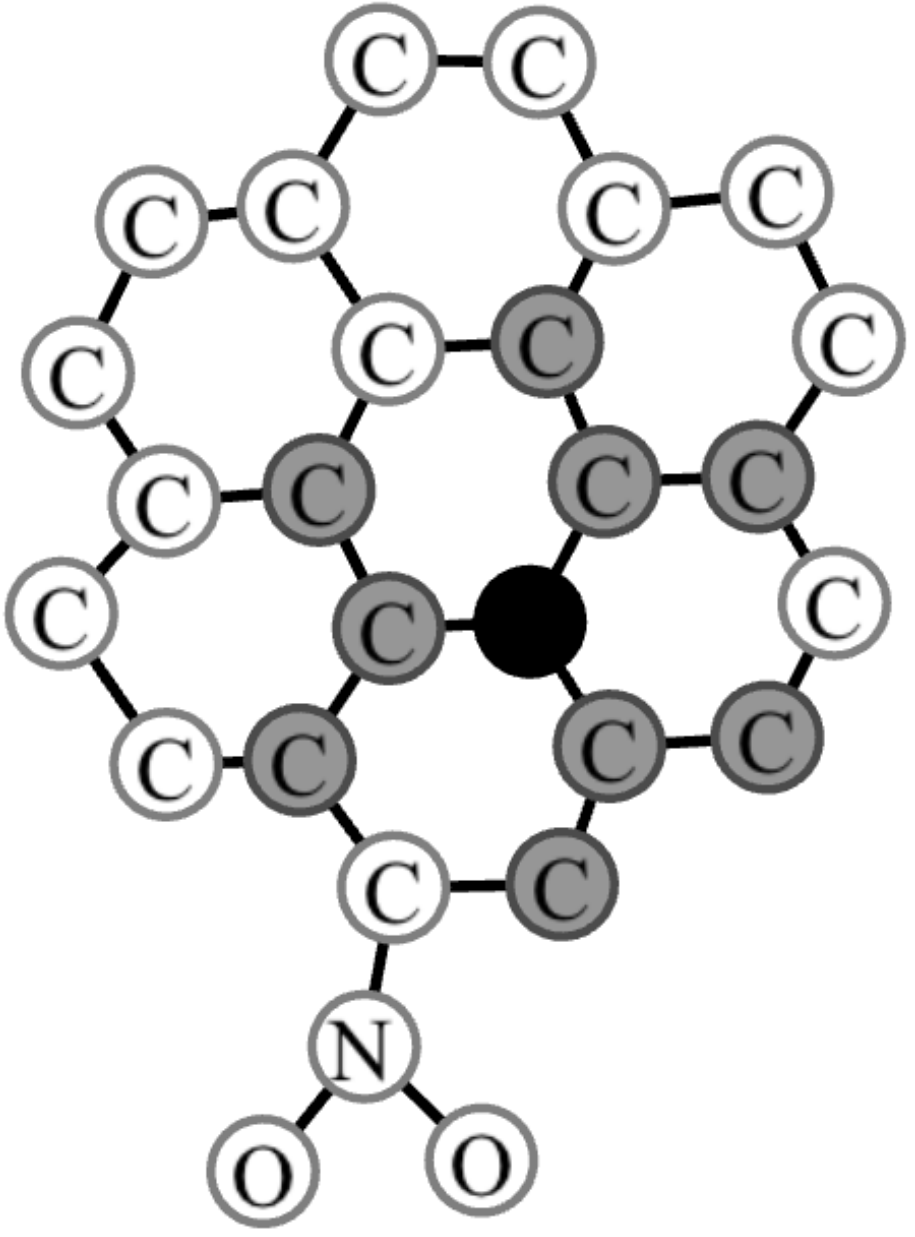} & \includegraphics[width=0.175\textwidth]{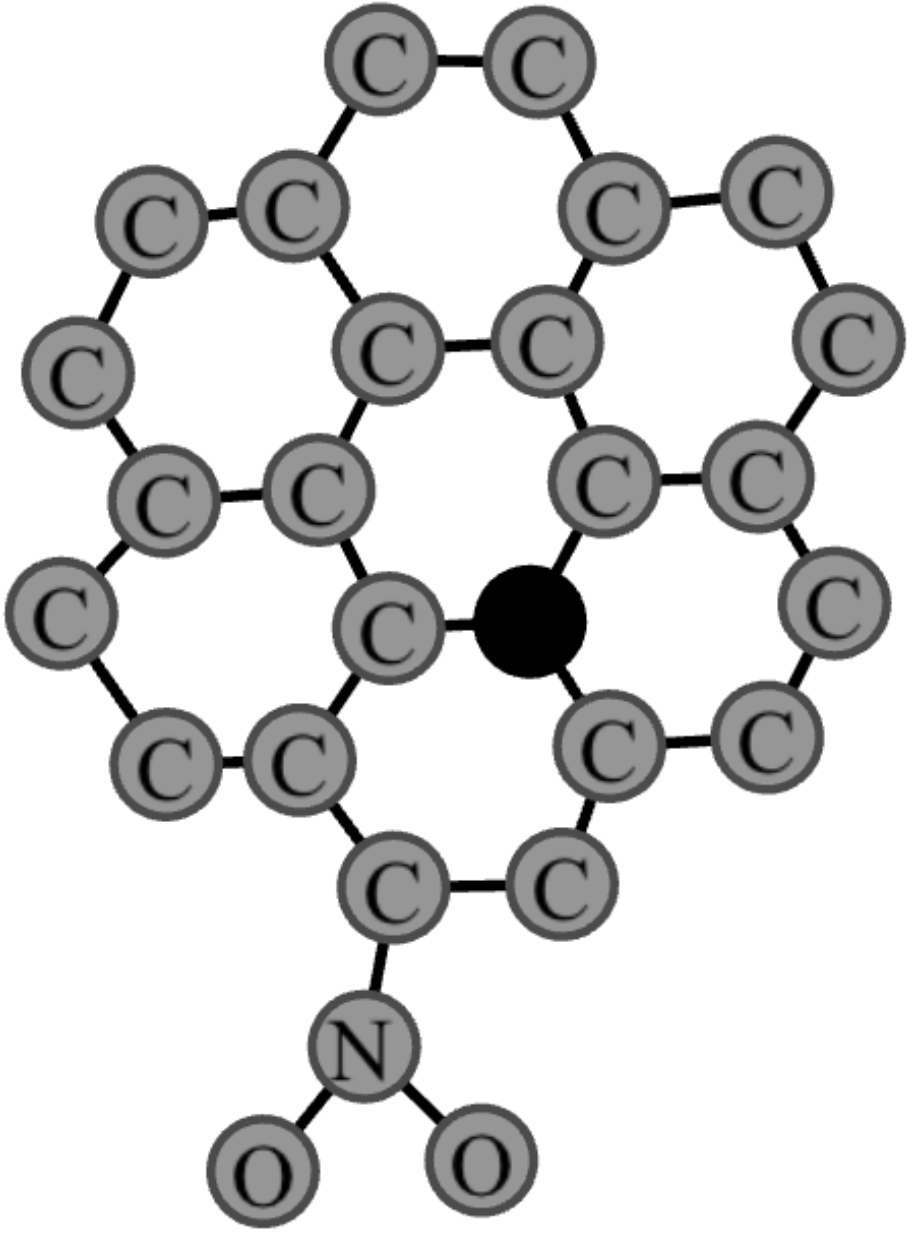} & \includegraphics[width=0.175\textwidth]{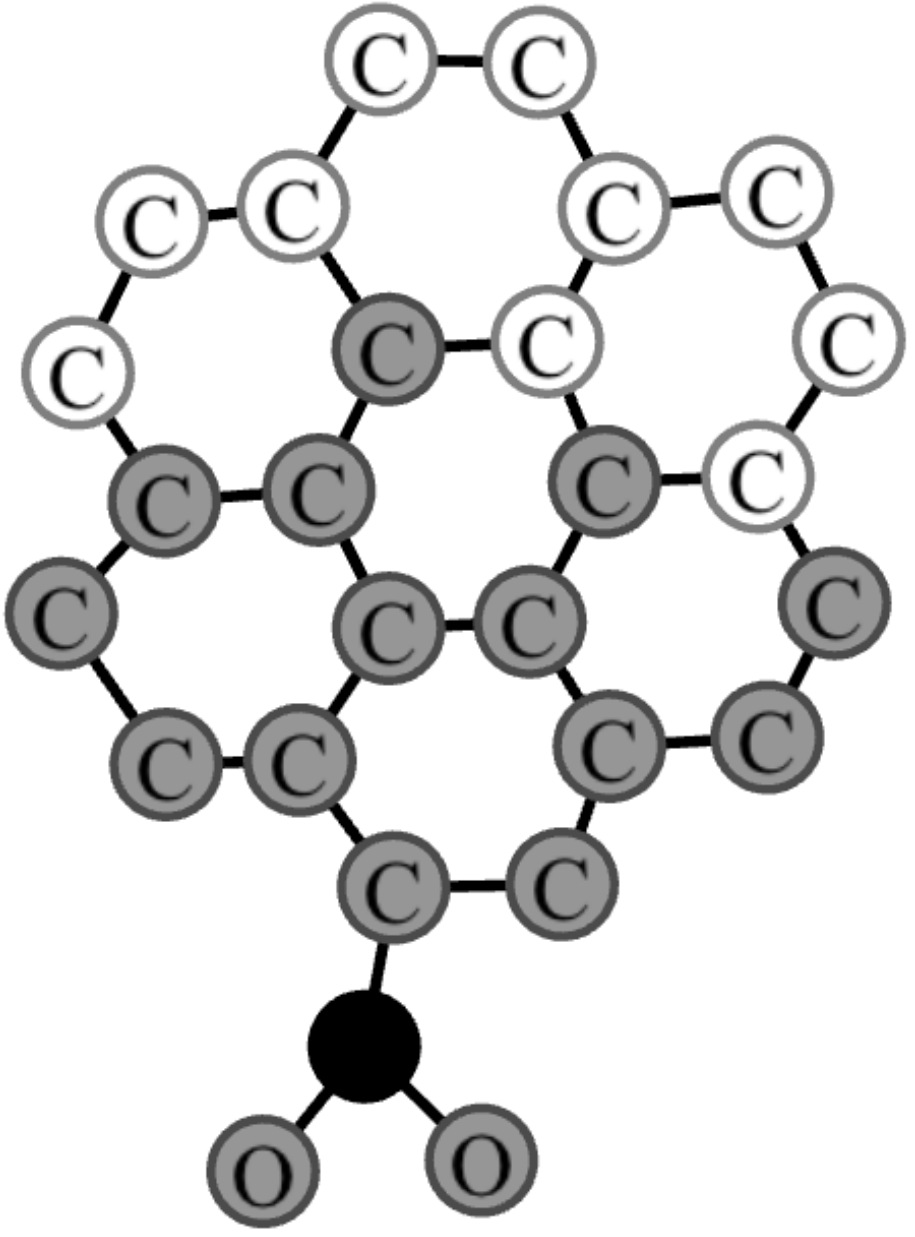}\tabularnewline
(a) & (b) & (c) & (d)\tabularnewline
\end{tabular}}
\par\end{centering}
\caption{\label{fig:effective-receptive-field} The receptive field of a neuron
in an Ego-CNN effectively enlar\textcolor{black}{ges at a deeper layer.
(a)-(c) Receptive fields of neurons at the 1st, 2nd, and 5th layer
corresponding to the same node. (d) Receptive field of another neuron
at the 5th layer that partially covers the graph. The difference in
the coverage reflects the position of the corresponding node.}}
\end{figure}

The neighborhood of a node $n$ at the $l$-th layer is recursively
defined as the stack-up of the latent representation of the node $n$
and the latent representations of the $K$ nearest neighbors of the
node $n$ in $\mathcal{G}$ at the $(l-1)$-th layer. In effect, a
neighborhood at the $l$-th layer is an $l$-hop ego network, as shown
in Figure \ref{fig:effective-receptive-field}. A neighborhood represents
a deterministic local region of $\mathcal{G}$, avoiding the dynamics
in the native extension of Patchy-San discussed above and allowing
the location independent patterns to be detected by the Ego-CNN filters.
As compared with the Message-Passing NNs (Eq. (\ref{eq:h-mpnn})),
the filters $\boldsymbol{W}^{(l,\cdot)}$'s parametrize the non-aggregated
representations of nodes, hence allowing the precise critical structures
to be backtracked via $\boldsymbol{W}^{(l,\cdot)}$'s layer-by-layer.
Note that Ego-CNNs are a generalization of a node embedding model
called 1-head-attention graph attention networks (1-head GATs) \cite{velickovic2018graph},
where the $\boldsymbol{W}^{(l,d)}$ in Eq. (\ref{eq:model-neighborhood})
is replaced by a rank-1 matrix $\boldsymbol{C}^{(l,d)}$. The 1-head
GATs were proposed for node classification problems. When it is applied
to graph learning tasks, requiring the $\boldsymbol{C}^{(l,d)}$ to
be a rank-1 matrix severely limits model capacity and leads to degraded
task performance. We will show this in Section \ref{sec:experiments}.

\textbf{Ego-Centric Layers.} Note that in Eq. (\ref{eq:model-neighborhood})
the $K$ nearest neighbors $Nbr(n,\cdot)$'s are determined from the
input $\mathcal{G}$ and remain the same across all layers. This allows
the receptive fields of neurons corresponding to the same node to
be exponentially enlarged (in number of nodes) at deeper layers, as
shown in Figure \ref{fig:effective-receptive-field}. Furthermore,
since each $\boldsymbol{H}_{Nbr(n,\cdot),:}^{(l-1)}$ in Eq. (\ref{eq:model-neighborhood})
already represents an embedding of an $(l-1)$-hop ego network centering
at a node neighboring $n$, the filters $\boldsymbol{W}^{(l,\cdot)}$'s
in the next layer, when scanning $\boldsymbol{E}^{(n,l)}$, can reuse
$\boldsymbol{H}_{Nbr(n,1),:}^{(l-1)},\cdots,\boldsymbol{H}_{Nbr(n,K),:}^{(l-1)}$
to efficiently detect patterns in the $l$-hop ego network centering
at node $n$. An Ego-CNN enjoys the exponentially increased efficiency
in detecting large-scale critical structures.

In practice, one should configure the number of layers $L$ (a hyperparameter)
according to the diameter of $\mathcal{G}$ to ensure that the critical
structures can be detected at the global scale. As large social networks
usually manifest the small-world property \cite{watts1998collective},
$L$ is not likely to be a very large number. In addition, one can
extend the Ego-CNN model described above in different ways. For example,
an Ego-CNN can have different numbers of filters/neurons at different
layers. One can also pair up Ego-CNN with an existing node embedding
model \cite{cai2018comprehensive} that takes into account node/edge
features to compute better $\boldsymbol{H}_{n,:}^{(0)}$ for each
node. In fact, Ego-CNN can take any kind of node embeddings as input,
as shown in the left of Figure \ref{fig:model-egocnn}. 

\subsection{Visualizing Critical Structures}

\label{sec:visualization} Since an Ego-CNN is jointly trained with
the task model (to detect task-specific critical structures), the
applicable visualization techniques may vary from task to task. Here,
we propose a general visualization technique based on the Transposed
Deconvolution \cite{zeiler2011adaptive} that works alongside any
task model. It consists of two steps: (1) we add an Attention layer
\cite{itti1998model} between the last Ego-Convolution layer and the
first layer of the task model to find the most important neighborhoods
at the deepest Ego-Convolution layer. (2) We then use the Transposed
Deconvolution to backtrack the nodes in $\mathcal{G}$ that are covered
by each of the important neighborhoods identified in Step 1. For more
details, please refer to Section 2 of the supplementary materials.

\begin{figure}
\centering{}\includegraphics[height=4cm]{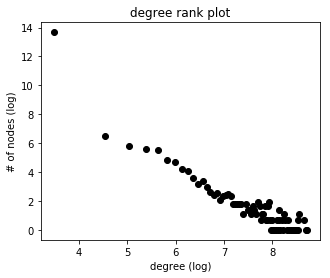}\caption{\label{fig:reddit-scale-free} The degree distribution of Reddit dataset
follows the power-law distribution.}
\end{figure}

\begin{table*}
\caption{\textcolor{black}{\label{tab:graph-classification-bioinformatic}10-Fold
CV test accuracy (\%) on bioinfomatic datasets. }}
\vspace{2mm}

\centering{}\resizebox{.7\textwidth}{!}{%
\begin{tabular*}{0.8\textwidth}{@{\extracolsep{\fill}}ccccc}
\toprule 
Dataset & MUTAG & PTC & PROTEINS & NCI1\tabularnewline
\midrule 
Size & 188 & 344 & 1113 & 4110\tabularnewline
Max \#node / \#class & 28 / 2 & 64 / 2 & 620 / 2 & 125 / 2\tabularnewline
\midrule
\midrule 
WL kernel & 82.1 & 57.0 & 73.0 & 82.2\tabularnewline
\midrule 
DGK & 82.7 & 57.3 & 71.7 & 62.5\tabularnewline
\midrule 
Subgraph2vec & 87.2 & 60.1 & 73.4 & 80.3\tabularnewline
\midrule 
MLG & 84.2 & 63.6 & \textbf{76.1} & 80.8\tabularnewline
\midrule 
Structure2vec & 88.3 & \textendash{} & \textendash{} & \textbf{83.7}\tabularnewline
\midrule 
DCNN & 67.0 & 56.6 & \textendash{} & 62.6\tabularnewline
\midrule 
Patchy-San & 92.6 & 60.0 & 75.9 & 78.6\tabularnewline
\midrule 
1-head-attention{\small{} GAT} & 81.0 & 57.0 & 72.5 & 74.3\tabularnewline
\midrule
Ego-CNN & \textbf{93.1} & \textbf{63.8} & 73.8 & 80.7\tabularnewline
\bottomrule
\end{tabular*}}
\end{table*}

We select the neighborhoods with attention scores higher than a predefined
threshold in Step 1 as the important ones. Note that the Attention
layer \cite{itti1998model} added in Step 1 does not need to be trained
with the Ego-CNN and task models. It can be efficiently trained after
the Ego-CNN is trained. To do so, we append the Attention layer and
a dense layer with a linear activation function (acting as a linear
task model) to the last Ego-Convolution layer of the trained Ego-CNN,
then we train the weights of the Attention and dense layers while
leaving the weights of the Ego-Convolution layers in the Ego-CNN fixed.
The linearity of the task model aligns the attention scores with the
importance. This post-visualization technique allows a model user
to quickly explore different network configurations for visualization.

\subsection{Efficiency and the Scale-Free Prior}

\label{sec:scale-free-regularizer}Given a graph with $N$ nodes and
$D$-dimensional embeddings of nodes, an Ego-CNN with $L$ Ego-Convolution
layers base on the top-$K$ neighbors and $D$ filters can embed a
graph in $O\left(N(K+1)LD^{2}\right)$ time. For each of the $L$
layers, the $l$-th layer takes $O(NK)$ to lookup and stack up the
$K$ neighbors' embeddings to generate all the $N$ receptive fields
of size $(K+1)\times D$, and it takes $O\left(N(K+1)D^{2}\right)$
to have $D$ filters scan through all the receptive fields. The Ego-CNN
is highly efficient as compare to existing graph embedding models.
Please see Table \ref{tab:graph-embedding-model-comparison} for more
details. 

\textbf{Scale-Free Regularizer. }Study \cite{li2005towards} shows
that the patterns in a large social network are usually scale-free\textemdash the
\textit{same patterns} can be observed at different zoom levels of
the network.\footnote{Interested readers may refer to \cite{kim2007box} for a formal definition
of a scale-free network, which is based on the fractals and box-covering
methods.} In practice, one may identify a scale-free network by checking if
the node degrees follow a power-law distribution. Figure \ref{fig:reddit-scale-free}
shows the degree distribution of the Reddit dataset, which is used
as one of the datasets in our experiment. The degree distribution
follows the power law.

The Ego-CNNs can be readily adapted to detect the scale-free patterns\footnote{For example, Kronecker graphs \cite{leskovec2010kronecker} is a special
case of the weight-tying Ego-CNN with filter number $D=1$. Interested
readers may refer to Section 4 of the supplementary for more details.}. Recall that the filters at the $l$-th layer detect the \textit{patterns}
of neighborhoods representing the $l$-hop ego networks. By regarding
the $1$-hop, $2$-hop, $\cdots$, $L$-hop ego-networks $\boldsymbol{E}^{(n,1)},\boldsymbol{E}^{(n,2)},\cdots,\boldsymbol{E}^{(n,L)}$
centering around the same node $n$ as different ``zoom levels''
of the graph, we can simply let an Ego-CNN detect the scale-free patterns
by tying the weights of filters $\boldsymbol{W}^{(1,d)},\boldsymbol{W}^{(2,d)},\cdots,\boldsymbol{W}^{(L,d)}$
for each $d$. When the input $\mathcal{G}$ is scale-free, this weight-tying
technique (a regularization) improves both the performance of the
task model and training efficiency.

\section{Experiments}

\label{sec:experiments}In this section, we conduct experiments using
real-world datasets to verify (i) Ego-CNNs can lead to comparable
task performance as compared to existing graph embedding approaches;
(ii) the visualization technique discussed in Section \ref{sec:visualization}
can output meaningful critical structures; and (iii) the scale-free
regularizer introduced in Section \ref{sec:scale-free-regularizer}
can detect the repeating patterns in a scale-free network. All experiments
run on a computer with 48-core Intel(R) Xeon(R) E5-2690 CPU, 64 GB
RAM, and NVidia Geforce GTX 1070 GPU. We use Tensorflow to implement
our methods. 

\subsection{Graph Classification}

\label{sec:exp-settings}We benchmark on both bioinformatic and social-network
datasets pre-processed by \cite{KKMMN2016}. In the bioinformatic
datasets, graphs are provided with node/edge labels and/or attributes,
while in the social network datasets, only pure graph structures are
given. We consider the task of graph classification. See DGK \cite{yanardag2015deep}
for more details about the task and benchmark datasets. We follow
DGK to set up the experiments and report the average test accuracy
using the 10-fold cross validation (CV). We compare the results Ego-CNN
with existing methods mentioned in Section \ref{sec:related-work}
and take the reported accuracy directly from their papers. 

\textbf{Generic Model Settings.} To demonstrate the broad applicability
of Ego-CNNs, the network architecture of our Ego-CNN implementation
remains the same for all datasets.  The architecture is composed
of 1 node embedding layer (Patchy-San with 128 filters and $K=10$)
and 5 Ego-Convolution layers (each with $D=128$ filters and $K=16$)
and 2 Dense layers (with 128 neurons for the first Dense layer) as
the task model before the output. We apply Dropout (with drop rate
$0.5$) and Batch Normalization to the input and Ego-Convolution layers
and train the network using the Adam algorithm with learning rate
$0.0001$. For selecting the $K$ neighbors, we exploit a heuristic
that prefers rare neighbors. We select the top $K$ with the least
frequent multiset labels in 1-WL labeling \cite{weisfeiler1968reduction}.
For nodes with less than $K$ neighbors, we simply use zero vectors
to represent non-existing neighbors.

The task accuracy are reported in Table \ref{tab:graph-classification-bioinformatic}
and Table \ref{tab:graph-classification-social-network}. Although
 having fixed architecture, the Ego-CNN is able to give comparable
task performance against the stat-of-the-art models (which all use
node/edge features) on the bioinformatic datasets. On the social network
datasets where the node/edge features are not available, the Ego-CNN
is able to outperform previous scalable work. In particular, the Ego-CNN
improves the performance of two closely related work, the single-layer
Patchy-San and 1-head-attention GAT, on most of the datasets. This
justifies that 1) detecting patterns at scales larger than just the
adjacent neighbors of each node and 2) allowing full-rank filters/kernels
in Eq. (\ref{eq:model-neighborhood}) are indeed beneficial. 

\subsection{Visualization of Critical Structures}

\textbf{Chemical Compounds.} To justify the usefulness of Ego-CNNs
in the cheminformatics problem shown in Figure \ref{fig:critical-struct-challenges},
we generate two compound datasets with critical structures at the
local scale (Alkanes vs. Alcohols) and at the global scale (Symmetric
vs. Asymmetric Isomers) in the ground truth, respectively. The structures
of compounds are generated under different compound size (number of
atoms) and vertex-orderings. 

\begin{table*}
\caption{\textcolor{black}{\label{tab:graph-classification-social-network}10-Fold
CV test accuracy (\%) on social network datasets. }}

\vspace{2mm}
\centering{}\resizebox{.7\textwidth}{!}{%
\begin{tabular*}{0.8\textwidth}{@{\extracolsep{\fill}}ccccc}
\toprule 
Dataset & IMDB (B) & IMDB (M) & REDDIT (B) & COLLAB\tabularnewline
\midrule 
Size & 1000 & 1000 & 2000 & 5000\tabularnewline
Max \#node / \#class & 270 / 2 & 176 / 3 & 3782 / 2 & 982 / 3\tabularnewline
\midrule
\midrule 
DGK & 67.0 & 44.6 & 78.0 & 73.0\tabularnewline
\midrule 
Patchy-San & 71.0 & 45.2 & 86.3 & 72.6\tabularnewline
\midrule 
1-head-attention{\small{} GAT} & 70.0 & \textendash{} & 78.8 & \textendash{}\tabularnewline
\midrule
Ego-CNN & \textbf{72.3} & \textbf{48.1} & \textbf{87.8} & \textbf{74.2}\tabularnewline
\bottomrule
\end{tabular*}}
\end{table*}
\begin{figure}
\begin{centering}
\begin{tabular}{cc}
\includegraphics[width=0.22\textwidth]{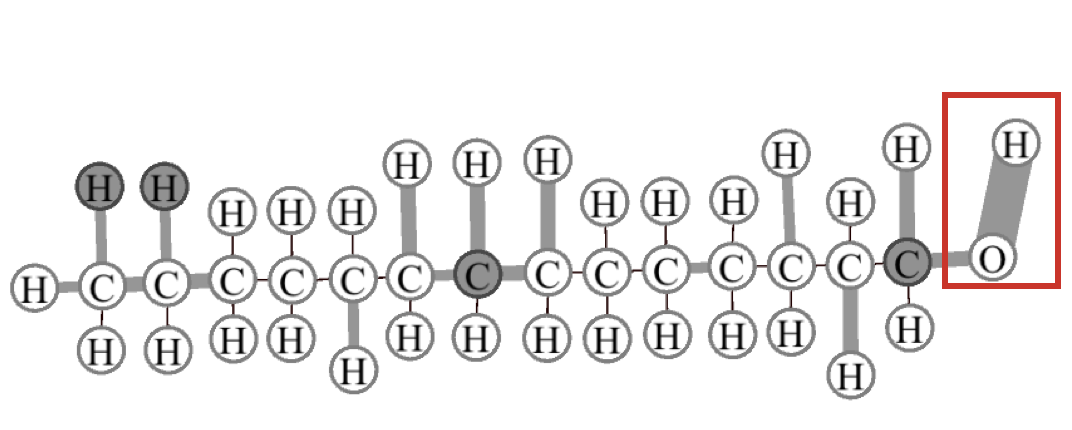} & \includegraphics[width=0.16\textwidth]{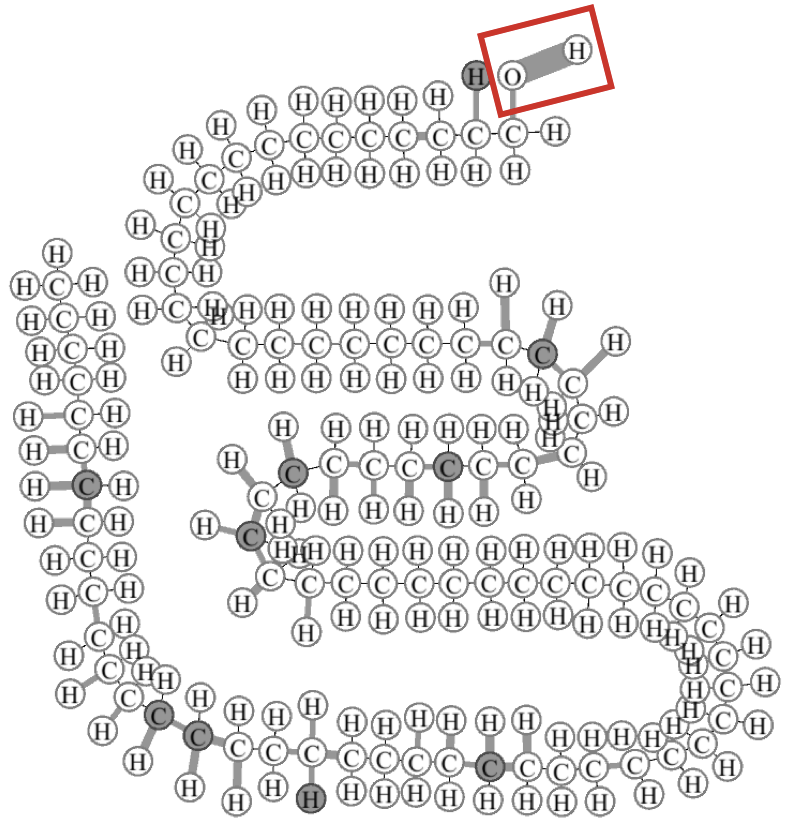}\tabularnewline
(a) C$_{14}$ H$_{29}$ OH & (b) C$_{82}$ H$_{165}$ OH\tabularnewline
 & \tabularnewline
\includegraphics[width=0.22\textwidth]{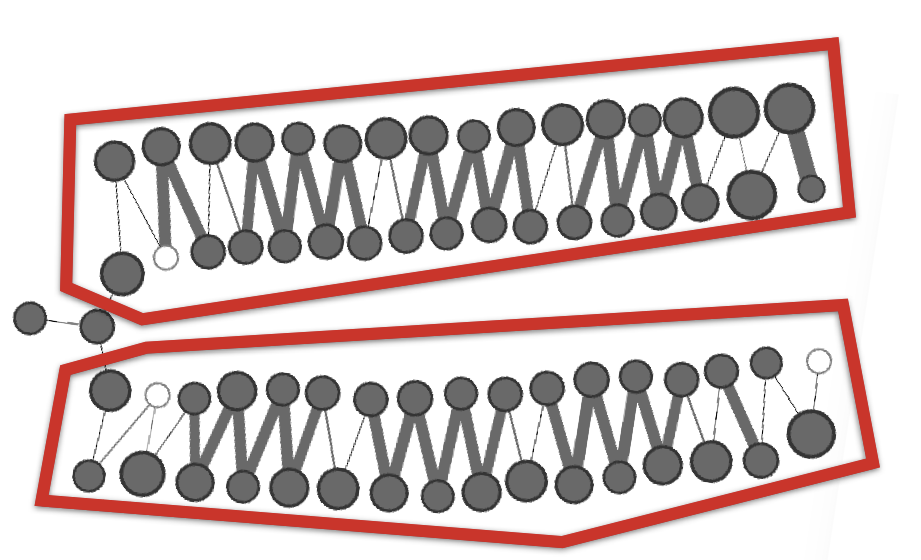} & \includegraphics[width=0.22\textwidth]{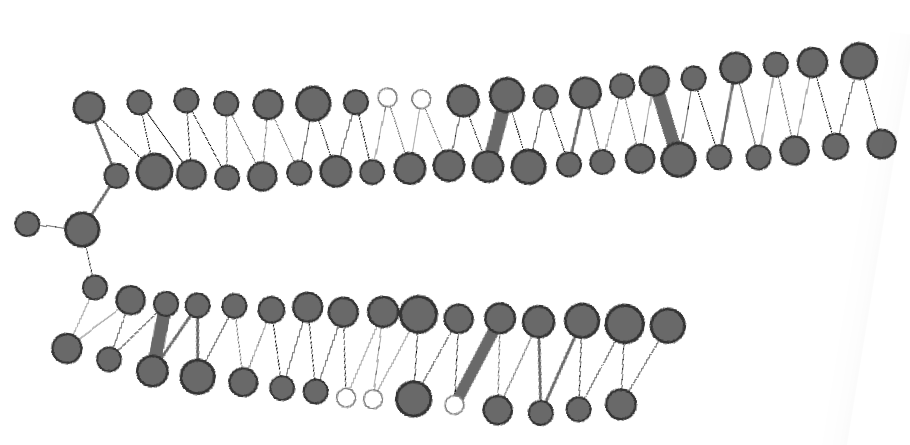}\tabularnewline
(c) Symmetric Isomer & (d) Asymmetric Isomer\tabularnewline
\end{tabular}
\par\end{centering}
\centering{}\caption{\label{fig:exp-cheminstry} Visualization of critical structures on
(a)(b) two Alcohol compounds for the task distinguishing Alcohol from
Alkane, and (c) a Symmetric Isomer and (d) an Asymmetric Isomer compounds
for the task classifying the types of Isomer. Critical structures
are colored in grey and the node/edge size is proportional to its
importance. The OH-base on Alcohols is always captured precisely and
considered critical. On Symmetric Isomers, the critical patterns are
roughly symmetric from the methyl branching node, which shows that
the Ego-CNN is able to learn to count from the branching node to see
if the structure is symmetric or not.}
\end{figure}

First, we test if an Ego-CNN considers OH-base as a critical structure
in the Alkanes vs. Alcohols dataset. With the post-visualization technique
introduced in Section \ref{sec:visualization}, we plot the detected
critical structures on two Alcohol examples in Figures \ref{fig:exp-cheminstry}(a)(b).
We find that the OH-base on Alcohols is always captured precisely
and considered as critical to distinguish Alcohols from Alkanes no
matter how large the compounds are.

For Symmetric Isomers like the one shown in Figure \ref{fig:exp-cheminstry}(c),
the Ego-CNN detects the symmetric hydrocarbon chains as critical structures
as we expected. An interesting observation is that the importance
of the nodes and edge in the detected critical structures are also
roughly symmetric to the methyl-base. This symmetry phenomenon can
also be observed in the critical structures of the Asymmetric Isomers,
as shown in Figure \ref{fig:exp-cheminstry}(d). We conjecture the
Ego-CNN learns to compare if the two long hydrocarbon chains (which
are branched from the methyl-base) are symmetric or not by starting
comparing the nodes and edges from the methyl-base all along to the
end of the hydrocarbon chains, which is similar to how people check
if a structure is symmetric.

\begin{figure}
\begin{centering}
\begin{tabular}{cc}
\includegraphics[width=0.25\textwidth]{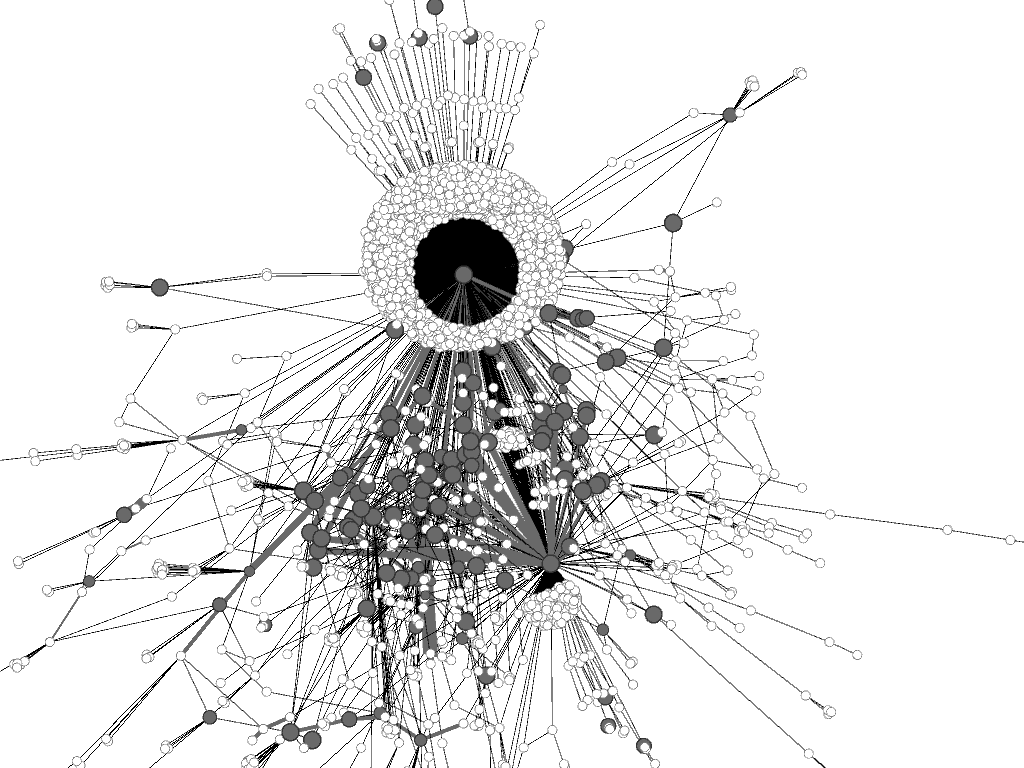} & \includegraphics[width=0.25\textwidth]{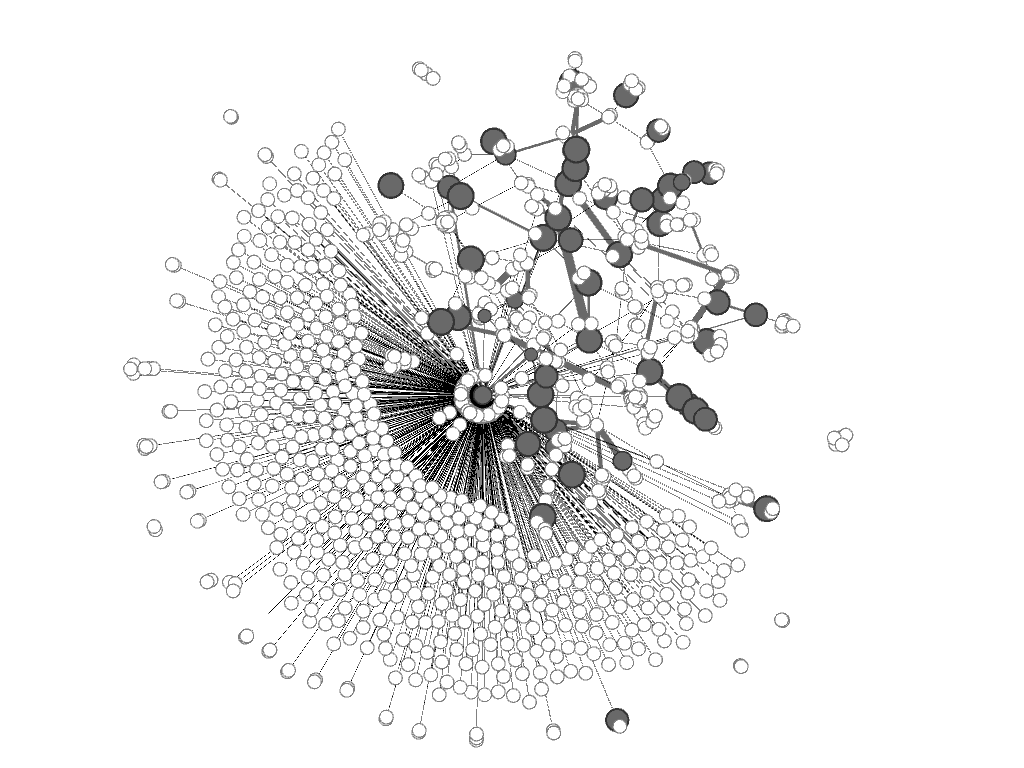}\tabularnewline
(a) Discussion-based thread. & (b) QA-based thread.\tabularnewline
\end{tabular}
\par\end{centering}
\caption{\label{fig:reddit-discussion} Visualization of critical structures
on Reddit dataset. The critical structures are colored in grey and
the node/edge size is proportional to its importance. The results
show that the variety of different opinions are the key to discriminant
discussion-based threads from QA-based threads.}
\end{figure}

\textbf{Social Interactions.} Without assuming prior knowledge, we
visualize the detected critical structures on graphs in the Reddit
dataset to see if they can help explain the task predictions. In Reddit
dataset, each graph represents a discussion thread. Each node represents
a user, and there is an edge if two users have been discussing with
each other. The task is to classify the discussion style of the thread
into either the discussion-based (e.g. threads under Atheism) or the
QA-based (e.g. under AskReddit).

Figure \ref{fig:reddit-discussion} shows the detected critical structures
(colored in grey with the node/edge size proportional to its importance).
For the discussion-based threads, the Ego-CNN tends to identify users
(nodes) that have many connections with other users. On the other
hand, many isolated nodes are identified as critical for the QA-based
threads. This suggests that the variety of different opinions, which
motivate following-up interactions between repliers in a tread, are
the key to discriminant discussion-based threads from QA-based threads.

\subsection{Scale-Free Regularizer}

Next, to verify the effectiveness of the scale-free regularizer proposed
in Section \ref{sec:scale-free-regularizer}. We compared 1 shallow
Ego-CNN (with 1 Ego-Convolution) and 2 deep Ego-CNNs (with 5 Ego-Convolution).
All networks are trained on the Reddit dataset with settings described
in Section \ref{sec:exp-settings}. Table \ref{tab:scale-free} shows
the results.

\begin{table}
\caption{\label{tab:scale-free} Ego-CNNs on Reddit dataset with the scale-free
prior.}
\vspace{2mm}{\small{}}%
\begin{tabular*}{1\columnwidth}{@{\extracolsep{\fill}}c>{\centering}p{1cm}>{\centering}p{1.75cm}>{\centering}p{1.75cm}}
\hline 
{\small{}Network architecture} & {\small{}Weight-Tying?} & {\small{}10-Fold CV Test Acc (\%)} & {\small{}\#Params}\tabularnewline
\hline 
\hline 
{\small{}1 Ego-Conv. layer} &  & {\small{}84.9} & \textbf{\small{}1.3M}\tabularnewline
\hline 
{\small{}5 Ego-Conv. layers} &  & {\small{}87.8} & {\small{}2.3M}\tabularnewline
\hline 
{\small{}5 Ego-Conv. layers} & {\small{}\Checkmark{}} & \textbf{\small{}88.4} & \textbf{\small{}1.3M}\tabularnewline
\hline 
\end{tabular*}
\end{table}

Without scale-free regularizer, the accuracy improves by 2.9\% at
the cost of 77\% more parameters. Tying the weights of the 5 Ego-Convolution
layers, the deep network uses roughly the same amount of parameters
as the shallow network but performs better than the network of the
same depth without weight-tying. This justifies that the proposed
scale-free regularizer can increase both the task performance and
training efficiency. 

Note, however, that the scale-free regularizer helps only when the
graphs are scale-free. When applied to graphs without scale-free properties
(e.g., chemical compounds), the scale-free regularizer leads to 2\%\textasciitilde 10\%
drop in test accuracy. For more details, please refer to Section 3
of the supplementary materials. This motivates a test like the one
shown in Figure \ref{fig:reddit-scale-free}\textemdash one should
verify if the target graphs indeed have scale-free properties before
applying the scale-free regularizer.

\section{Conclusions}

\label{sec:conclusions}
We propose Ego-CNNs that employ the Ego-Convolutions to detect invariant patterns among ego networks, and use the ego-centric way to stack up layers to allows to exponentially cover more nodes. The Ego-CNNs work nicely with common visualization techniques to illustrate the detected structures. Investigating the critical structures may help explaining the reasons behind task predictions and/or discovery of new knowledge, which is important to many fields
such as the bioinformatics, cheminformatics, and social network analysis. As our future work, we will study how to further improve the time/space efficiency of an Ego-CNN.
A neighborhood of a node at a deep layer may overlap with that of another node at the same layer. Therefore, instead of letting a filter scan through all of the neighborhood embeddings at a layer, it might be acceptable to skip some neighborhoods. This can reduce embedding dimensions (space) and speed up computation.

\section{Acknowledgments}
This work is supported by the MOST Joint Research Center for AI Technology and All Vista Healthcare, Taiwan (MOST 108-2634-F-007-003-). We also thank the anonymous reviewers for their insightful feedbacks. 

\bibliographystyle{icml2019}
\bibliography{egocnn}

\begin{thebibliography}{27}
\providecommand{\natexlab}[1]{#1}
\providecommand{\url}[1]{\texttt{#1}}
\expandafter\ifx\csname urlstyle\endcsname\relax
  \providecommand{\doi}[1]{doi: #1}\else
  \providecommand{\doi}{doi: \begingroup \urlstyle{rm}\Url}\fi

\bibitem[Atwood \& Towsley(2016)Atwood and Towsley]{atwood2016diffusion}
Atwood, J. and Towsley, D.
\newblock Diffusion-convolutional neural networks.
\newblock In \emph{Proceedings of NIPS}, 2016.

\bibitem[Bruna et~al.(2013)Bruna, Zaremba, Szlam, and LeCun]{bruna2013spectral}
Bruna, J., Zaremba, W., Szlam, A., and LeCun, Y.
\newblock Spectral networks and locally connected networks on graphs.
\newblock In \emph{Proceedings of ICLR}, 2013.

\bibitem[Cai et~al.(2018)Cai, Zheng, and Chang]{cai2018comprehensive}
Cai, H., Zheng, V.~W., and Chang, K.
\newblock A comprehensive survey of graph embedding: problems, techniques and
  applications.
\newblock \emph{IEEE Transactions on Knowledge and Data Engineering}, 2018.

\bibitem[Cook(1971)]{cook1971complexity}
Cook, S.~A.
\newblock The complexity of theorem-proving procedures.
\newblock In \emph{Proceedings of the third annual ACM symposium on Theory of
  Computing}. ACM, 1971.

\bibitem[Dai et~al.(2016)Dai, Dai, and Song]{dai2016discriminative}
Dai, H., Dai, B., and Song, L.
\newblock Discriminative embeddings of latent variable models for structured
  data.
\newblock In \emph{Proceedings of ICML}, 2016.

\bibitem[Defferrard et~al.(2016)Defferrard, Bresson, and
  Vandergheynst]{defferrard2016convolutional}
Defferrard, M., Bresson, X., and Vandergheynst, P.
\newblock Convolutional neural networks on graphs with fast localized spectral
  filtering.
\newblock In \emph{Proceedings of NIPS}, 2016.

\bibitem[Duvenaud et~al.(2015)Duvenaud, Maclaurin, Iparraguirre, Bombarell,
  Hirzel, Aspuru-Guzik, and Adams]{duvenaud2015convolutional}
Duvenaud, D.~K., Maclaurin, D., Iparraguirre, J., Bombarell, R., Hirzel, T.,
  Aspuru-Guzik, A., and Adams, R.~P.
\newblock Convolutional networks on graphs for learning molecular fingerprints.
\newblock In \emph{Proceedings of NIPS}, 2015.

\bibitem[Gilmer et~al.(2017)Gilmer, Schoenholz, Riley, Vinyals, and
  Dahl]{gilmer2017neural}
Gilmer, J., Schoenholz, S.~S., Riley, P.~F., Vinyals, O., and Dahl, G.~E.
\newblock Neural message passing for quantum chemistry.
\newblock In \emph{Proceedings of ICML}, 2017.

\bibitem[Itti et~al.(1998)Itti, Koch, and Niebur]{itti1998model}
Itti, L., Koch, C., and Niebur, E.
\newblock A model of saliency-based visual attention for rapid scene analysis.
\newblock \emph{IEEE Transactions on Pattern Analysis and Machine
  Intelligence}, 20\penalty0 (11):\penalty0 1254--1259, 1998.

\bibitem[Kersting et~al.(2016)Kersting, Kriege, Morris, Mutzel, and
  Neumann]{KKMMN2016}
Kersting, K., Kriege, N.~M., Morris, C., Mutzel, P., and Neumann, M.
\newblock Benchmark data sets for graph kernels, 2016.
\newblock URL \url{http://graphkernels.cs.tu-dortmund.de}.

\bibitem[Kim et~al.(2007)Kim, Goh, Kahng, and Kim]{kim2007box}
Kim, J., Goh, K.-I., Kahng, B., and Kim, D.
\newblock A box-covering algorithm for fractal scaling in scale-free networks.
\newblock \emph{Chaos: An Interdisciplinary Journal of Nonlinear Science},
  2007.

\bibitem[Kipf \& Welling(2017)Kipf and Welling]{kipf2016semi}
Kipf, T.~N. and Welling, M.
\newblock Semi-supervised classification with graph convolutional networks.
\newblock In \emph{Proceedings of ICLR}, 2017.

\bibitem[Kondor \& Pan(2016)Kondor and Pan]{kondor2016multiscale}
Kondor, R. and Pan, H.
\newblock The multiscale laplacian graph kernel.
\newblock In \emph{Proceedings of NIPS}, 2016.

\bibitem[Leskovec et~al.(2010)Leskovec, Chakrabarti, Kleinberg, Faloutsos, and
  Ghahramani]{leskovec2010kronecker}
Leskovec, J., Chakrabarti, D., Kleinberg, J., Faloutsos, C., and Ghahramani, Z.
\newblock Kronecker graphs: An approach to modeling networks.
\newblock \emph{Journal of Machine Learning Research}, 11\penalty0
  (Feb):\penalty0 985--1042, 2010.

\bibitem[Li et~al.(2005)Li, Alderson, Doyle, and Willinger]{li2005towards}
Li, L., Alderson, D., Doyle, J.~C., and Willinger, W.
\newblock Towards a theory of scale-free graphs: Definition, properties, and
  implications.
\newblock \emph{Internet Mathematics}, pp.\  431--523, 2005.

\bibitem[Li et~al.(2016)Li, Tarlow, Brockschmidt, and Zemel]{li2016gated}
Li, Y., Tarlow, D., Brockschmidt, M., and Zemel, R.
\newblock Gated graph sequence neural networks.
\newblock In \emph{Proceedings of ICLR}, 2016.

\bibitem[Mikolov et~al.(2013)Mikolov, Chen, Corrado, and
  Dean]{mikolov2013efficient}
Mikolov, T., Chen, K., Corrado, G., and Dean, J.
\newblock Efficient estimation of word representations in vector space.
\newblock 2013.

\bibitem[Narayanan et~al.(2016)Narayanan, Chandramohan, Chen, Liu, and
  Saminathan]{narayanan2016subgraph2vec}
Narayanan, A., Chandramohan, M., Chen, L., Liu, Y., and Saminathan, S.
\newblock subgraph2vec: Learning distributed representations of rooted
  sub-graphs from large graphs.
\newblock \emph{In Workshop on Mining and Learning with Graphs}, 2016.

\bibitem[Niepert et~al.(2016)Niepert, Ahmed, and Kutzkov]{niepert2016learning}
Niepert, M., Ahmed, M., and Kutzkov, K.
\newblock Learning convolutional neural networks for graphs.
\newblock In \emph{Proceedings of ICML}, 2016.

\bibitem[Pham et~al.(2017)Pham, Tran, Phung, and Venkatesh]{pham2017column}
Pham, T., Tran, T., Phung, D.~Q., and Venkatesh, S.
\newblock Column networks for collective classification.
\newblock In \emph{Proceedings of AAAI}, 2017.

\bibitem[Shervashidze et~al.(2011)Shervashidze, Schweitzer, Leeuwen, Mehlhorn,
  and Borgwardt]{shervashidze2011weisfeiler}
Shervashidze, N., Schweitzer, P., Leeuwen, E. J.~v., Mehlhorn, K., and
  Borgwardt, K.~M.
\newblock Weisfeiler-lehman graph kernels.
\newblock \emph{JMLR}, 12\penalty0 (Sep):\penalty0 2539--2561, 2011.

\bibitem[Velickovic et~al.(2018)Velickovic, Cucurull, Casanova, Romero, Lio,
  and Bengio]{velickovic2018graph}
Velickovic, P., Cucurull, G., Casanova, A., Romero, A., Lio, P., and Bengio, Y.
\newblock Graph attention networks.
\newblock In \emph{Proceedings of ICLR}, 2018.

\bibitem[Watts \& Strogatz(1998)Watts and Strogatz]{watts1998collective}
Watts, D.~J. and Strogatz, S.~H.
\newblock Collective dynamics of small-worldnetworks.
\newblock \emph{Nature}, 393\penalty0 (6684):\penalty0 440, 1998.

\bibitem[Weisfeiler \& Lehman(1968)Weisfeiler and
  Lehman]{weisfeiler1968reduction}
Weisfeiler, B. and Lehman, A.
\newblock A reduction of a graph to a canonical form and an algebra arising
  during this reduction.
\newblock \emph{Nauchno-Technicheskaya Informatsia}, 2\penalty0 (9):\penalty0
  12--16, 1968.

\bibitem[Yanardag \& Vishwanathan(2015)Yanardag and
  Vishwanathan]{yanardag2015deep}
Yanardag, P. and Vishwanathan, S.
\newblock Deep graph kernels.
\newblock In \emph{Proceedings of SIGKDD}. ACM, 2015.

\bibitem[Ying et~al.(2018)Ying, You, Morris, Ren, Hamilton, and
  Leskovec]{ying2018hierarchical}
Ying, Z., You, J., Morris, C., Ren, X., Hamilton, W., and Leskovec, J.
\newblock Hierarchical graph representation learning with differentiable
  pooling.
\newblock In \emph{Proceedings of NIPS}, pp.\  4805--4815, 2018.

\bibitem[Zeiler et~al.(2011)Zeiler, Taylor, and Fergus]{zeiler2011adaptive}
Zeiler, M.~D., Taylor, G.~W., and Fergus, R.
\newblock Adaptive deconvolutional networks for mid and high level feature
  learning.
\newblock In \emph{Proceedings of ICCV}. IEEE, 2011.

\end{thebibliography}

\makeatother
\twocolumn[
\icmltitle{Distributed, Egocentric Representations of Graphs for Detecting Critical Structures: Supplementary Metarials}

\icmlkeywords{Graph Embeddings, Convolutional Neural Networks} 
\vskip 0.3in]

\setcounter{section}{0}
\section{Further Related Work}

\label{sec:related-work}Here, we give an in-depth review of existing
graph embedding models.

\textbf{Graph Kernels.}  The Weisfeiler-Lehman kernel \cite{shervashidze2011weisfeiler}
grows the coverage of each node by collecting information from neighbors,
which is conceptually similar to our method, but differs from our
model in that WL kernel collects only node labels, while our method
collects the complete labeled neighborhood graphs from neighbors.
Deep Graph Kernels \cite{yanardag2015deep} and Subgraph2vec \cite{narayanan2016subgraph2vec},
which are inspired by word2vec \cite{mikolov2013efficient}, embed
the graph structure by predicting neighbors' structures given a node.
Multiscale Laplacian Graph Kernels \cite{kondor2016multiscale} compare
graphs at multiple scales by recursively comparing graphs based on
the comparison of subgraphs, which takes $O(LN^{5})$ where $L$ represents
the number of comparing scales and is inefficient. All the above graph
kernels have a common drawback in that the embeddings are generated
in a unsupervised manner. The critical structure cannot be jointly
detected at the generation of embeddings.

\textbf{Graphical Models.} Assuming that the edges of a graph express
the conditional dependency between random variables (nodes), Structure2vec
\cite{dai2016discriminative} introduces a novel layer that makes
the optimization procedures of approximated inference directly trainable
by SGD. It is efficient on large graphs with time complexity linear
to number of nodes. However, it's weak on identifying critical structures
because the approximated inference makes too much simplification on
the graph structure. For example, the mean-field approximation assumes
that variables are independent with each other. As a result, Structure2vec
can only identify critical structures of very simple shape.

\textbf{Convolution-based Methods.} Recently, many work are proposed
to embed graphs by borrowing the concept of CNN. Figure \ref{fig:cnn-filters}
summarizes the definitions of filters and neighborhoods in these work.
The Spatial Graph Convolutional Network (GCN) was proposed by \cite{bruna2013spectral}.
The design of Spatial GCN (Figure \ref{fig:cnn-filters}(a)) is very
different from other convolution-based methods (and ours) since its
goal is to perform hierarchical clustering of nodes. A neighborhood
is defined as a cluster. However, the filter is not aim to scan for
local patterns but to learn the connectivity of all clusters. This
means each filter is of size $O(N^{2})$ if there are $N$ clusters.
Also, a filter requires the global-scale information, i.e. the features
of all clusters to train, so it's very inefficient on large graphs.
Hence, in the same paper, \cite{bruna2013spectral} proposed another
version, the Spectrum GCN, to perform hierarchical clustering in the
spectrum domain. The computation of Spectrum GCN is later improved
by \cite{defferrard2016convolutional}. However, the major drawback
is that the graph spectrum is weak at identifying structures as it
is only interpretable for very special graph families (e.g., complete
graphs and star-like trees). A recent variant of Spectrum GCN \cite{kipf2016semi}
uses the filters to detect the propagated feature of each node (equivalent
to weighted-sum its neighbors' features). This variant also cannot
detect the critical structures precisely.

\begin{figure*}
\begin{centering}
\begin{tabular}{ccccc}
\includegraphics[width=0.1\textwidth]{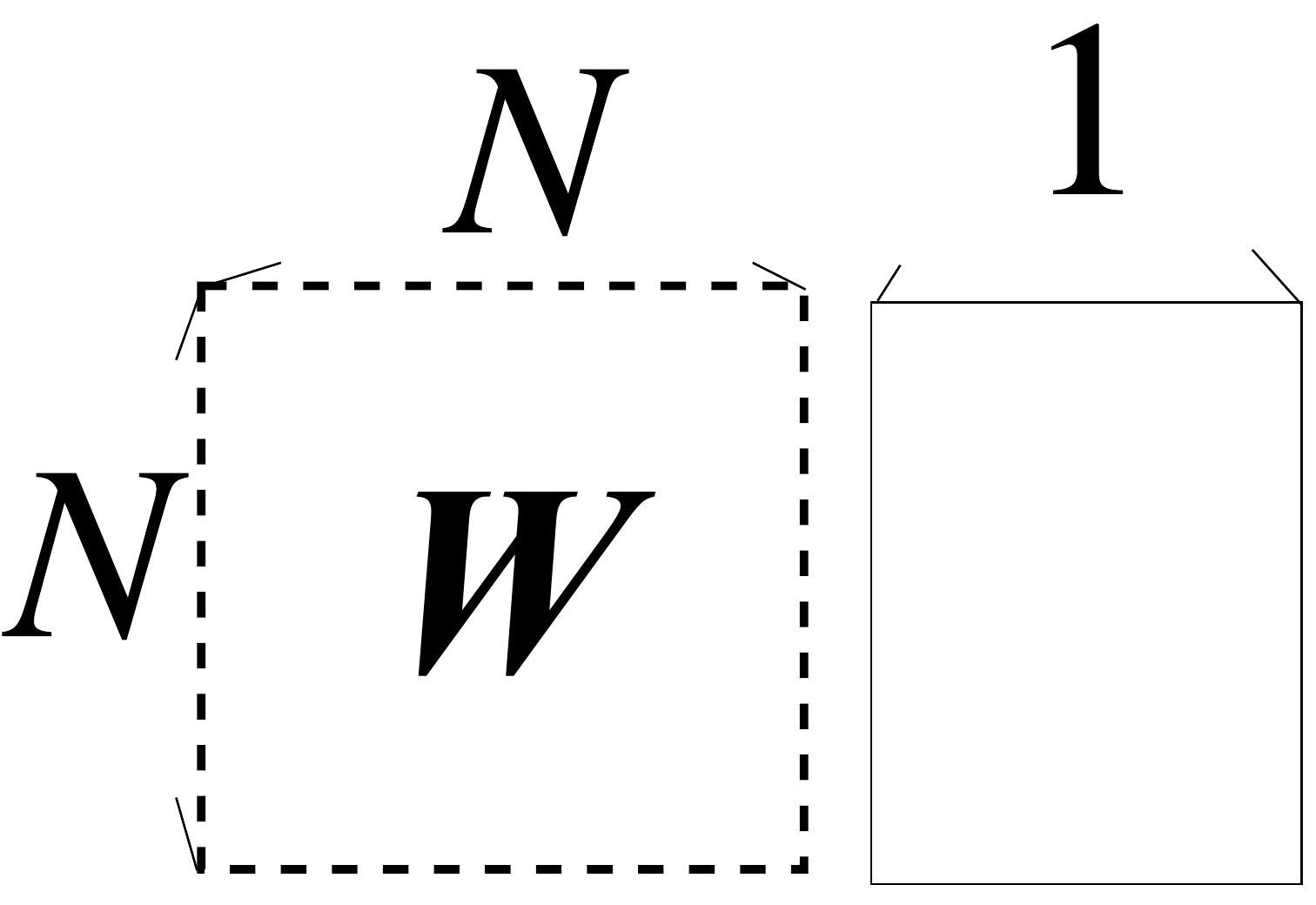} & \includegraphics[width=0.095\textwidth]{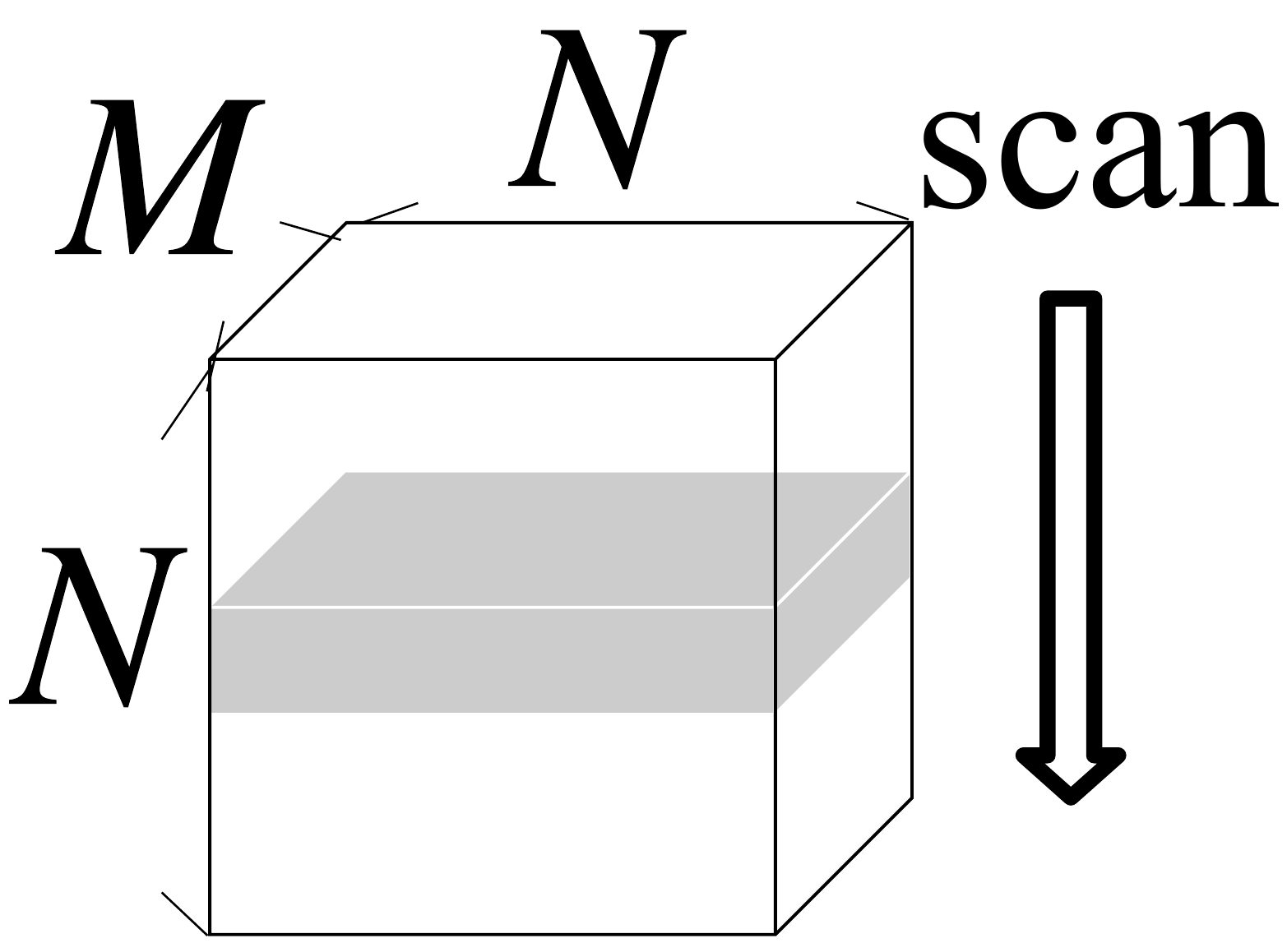} & \includegraphics[width=0.095\textwidth]{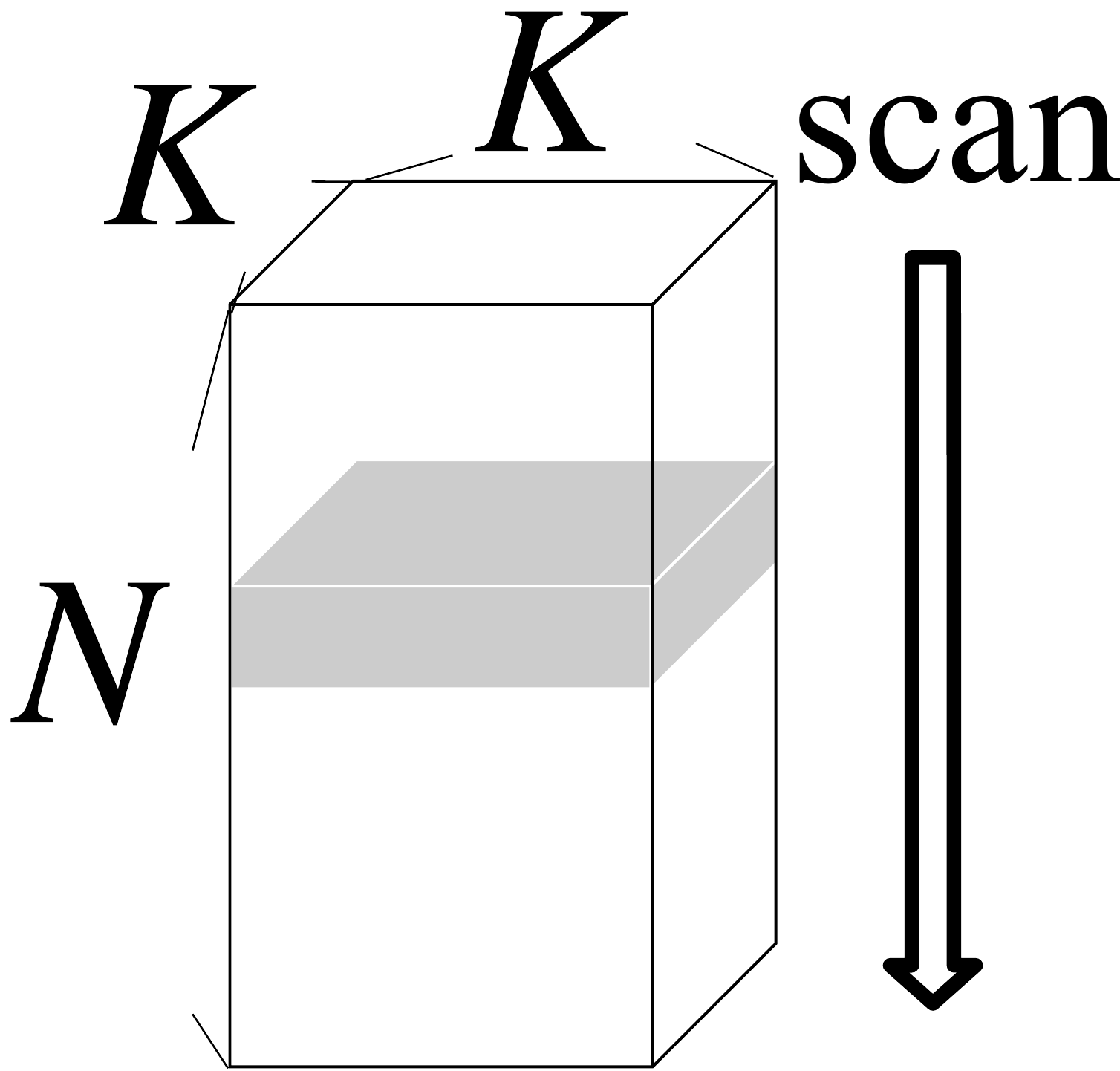} & \includegraphics[width=0.085\textwidth]{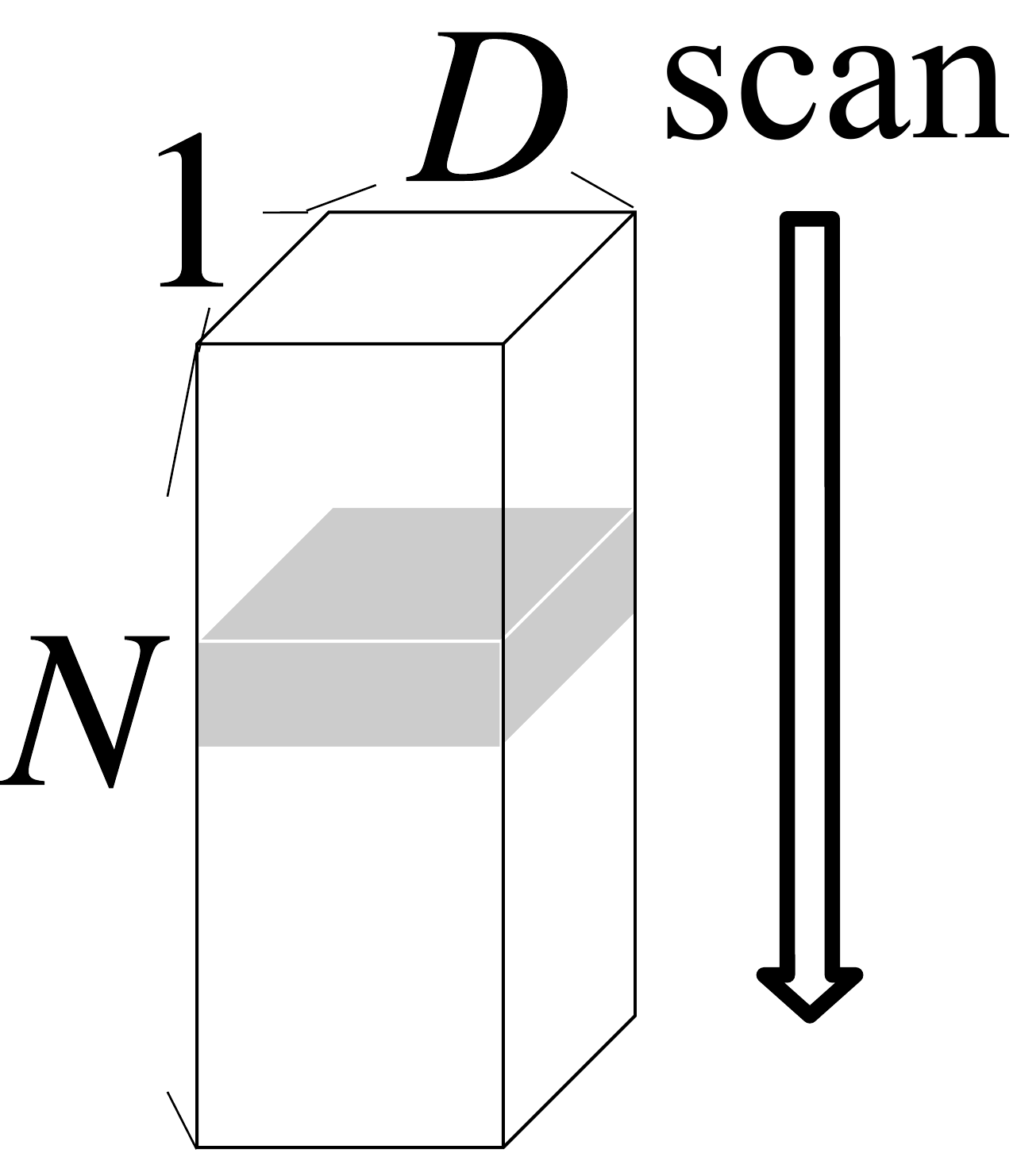} & \includegraphics[width=0.085\textwidth]{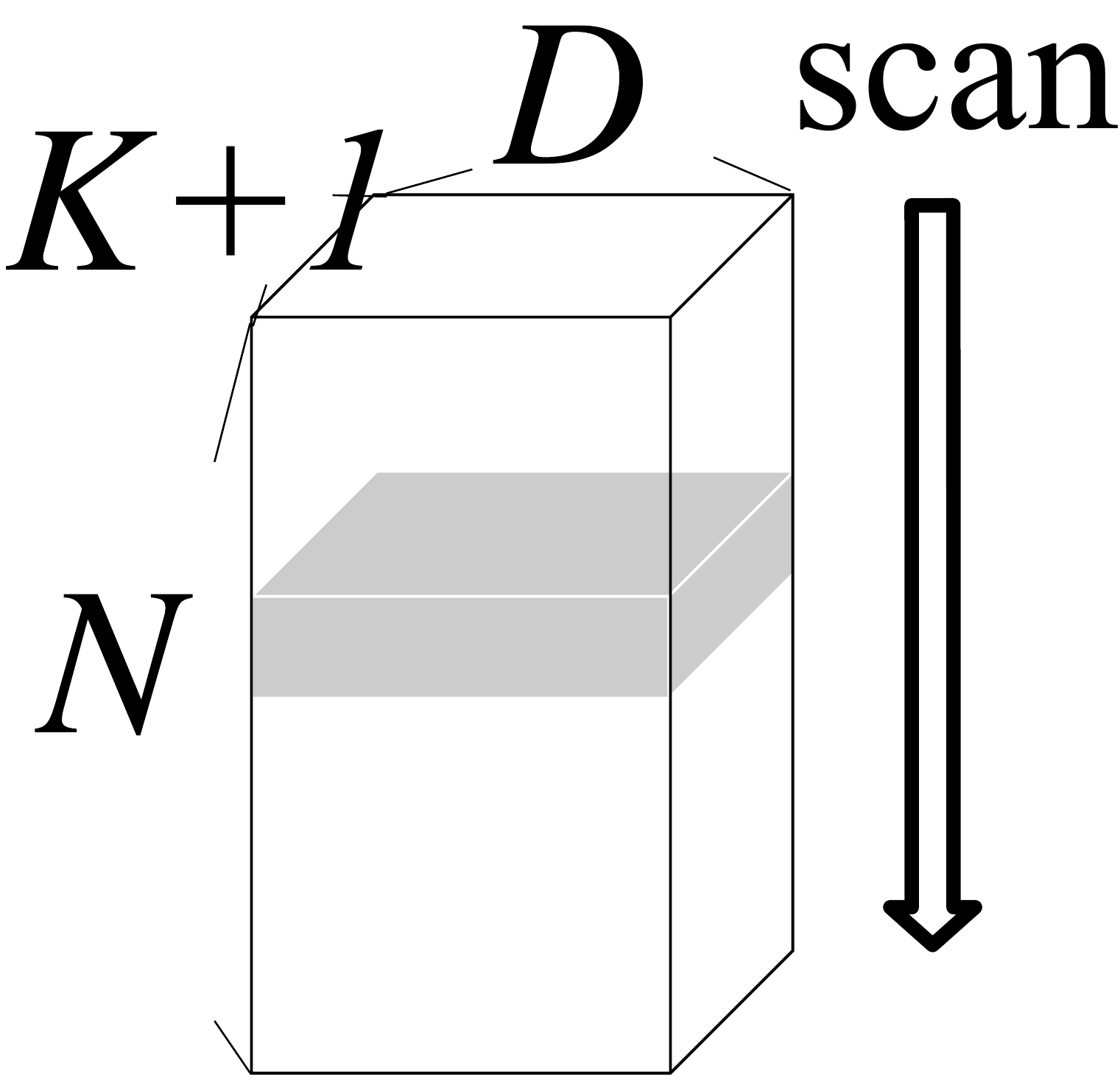}\tabularnewline
(a) & (b) & (c) & (d) & (e)\tabularnewline
\end{tabular}
\par\end{centering}
\caption{\label{fig:cnn-filters}Filters and neighborhood in (a) Spatial GCN
\cite{bruna2013spectral}. A filter $\boldsymbol{W}$ is a sparse
matrix, which is not aimed to detect local patterns, but to learn
the connectivity of clusters. (b) DCNN \cite{atwood2016diffusion}
scans through the $M\times N$ diffusion matrix of each node. (c)
Patchy-San \cite{niepert2016learning} scans the $K\times K$ adjacency
matrix of local neighborhood of each node. (d) Neural Fingerprints
\cite{duvenaud2015convolutional} that scans the $d$-dimensional
approximated neighborhood (specially, the summation of neighbors'
embeddings) of each node. (e) Ego-Convolution scans $(K+1)\times D$
egocentric neighborhoods of each node.}
\end{figure*}

Diffusion Convolutional Neural Networks (DCNN) \cite{atwood2016diffusion}
embed the graph by detecting patterns in the diffusion of each node.
The neighborhood is defined as the diffusion, which are paths starting
from a node to other nodes in $m$ hops. The diffusion can be represented
by an $M\times N$ diffusion matrix $\boldsymbol{D}$, where each
element $D_{m,n}$ indicates if the current node connects to a node
$n$ in $m$ hops. Filters in a DCNN scan through each node's diffusion
matrix. So, DCNN can detect useful diffusion patterns. But it cannot
detect critical structures since the diffusion patterns cannot precisely
describe the location and the shape of structures. DCNN reported impressive
results on the node classification. But it is inefficient on graph
classification tasks since their notion of neighborhood is at the
global-scale, which takes $O(MN^{2})$ to embed a graph with $N$
nodes. Also, their definition of neighborhood makes the embedding
model shallow\textemdash there is only one single layer in a DCNN.

Patchy-San \cite{niepert2016learning} uses filters to detects patterns
in the adjacency matrix of the $K$ nearest neighbors of each node.
The neighborhood of a node $n$ is defined as the $K\times K$ adjacency
matrix $\boldsymbol{A}^{(n)}$ of the $K$ nearest neighbors of the
node. Filters $\boldsymbol{W}^{(d)}\in\mathbb{R}^{K\times K}$, $d=1,2,\cdots,D$,
scan through the adjacency matrix of each node to generate the graph
embedding $\boldsymbol{H}\in\mathbb{R}^{N\times D}$, where ${\displaystyle H_{n,d}=\sigma(\bar{\boldsymbol{A}}^{(n)}\varoast\boldsymbol{W}^{(d)}+b_{d}})$
is the output of an activation function $\sigma$, $b_{d}$ is the
bias term, and $\varoast$ is the Frobenius inner product defined
as $\boldsymbol{X}\varoast\boldsymbol{Y}={\displaystyle \sum_{i,j}X_{i,j}Y_{i,j}}$.
Note that a filter scans through $\bar{\boldsymbol{A}}^{(n)}$, the
normalized version of $\boldsymbol{A}^{(n)}$ \cite{niepert2016learning},
in order to be invariant under different vertex permutations. Patchy-San
can detect precise structures (via the filters). However, to detect
critical structures at the global level, each $\bar{\boldsymbol{A}}^{(n)}$
needs to have the size of $N\times N$, making the filters $\boldsymbol{W}^{(d)}\in\mathbb{R}^{N\times N}$
hard to learn. There is no discussion on how to generalize the local
neighborhood defined by Patchy-San at a deep layer. 

The Message-Passing NNs \cite{duvenaud2015convolutional,li2016gated,pham2017column,gilmer2017neural,velickovic2018graph,ying2018hierarchical}
scan through the approximated neighborhoods of each node and supports
multiple layers. At each layer $l$, the neighborhood of a node is
defined as a $D$-dimensional vector representing the aggregation
(e.g., summation \cite{duvenaud2015convolutional}) of the $D$-dimensional
hidden representation at layer $l-1$ of the $l$-hop neighbors. The
summation avoids the vertex-ordering problem of the adjacency matrix
in Patchy-San. On the other hand, the Message-Passing NNs loses the
ability of detecting precise critical structures.

Note that a Message-Passing NN, called the 1-head-attention graph
attention network (1-head GAT) \cite{velickovic2018graph}, is a special
case of Ego-CNN where the $\boldsymbol{W}^{(l,d)}$ in Eq. (3) in
the main paper is replaced by a rank-1 matrix $\boldsymbol{C}^{(l,d)}$.
Specifically, it models the $d$-th dimension of the embedding of
node $n$ at the $l$-th layer as $H_{n,d}^{(l)}=\sigma\left(\left[\boldsymbol{H}_{Nbr(n,1),:}^{(l-1)},\cdots,\boldsymbol{H}_{Nbr(n,K),:}^{(l-1)}\right]^{\top}\circledast\boldsymbol{C}^{(l,d)}\right)$,
where $\boldsymbol{C}^{(l,d)}=\boldsymbol{\alpha}\boldsymbol{W}_{d,:}$
is the outer-product of the edge importance vector $\boldsymbol{\alpha}\in\mathbb{R}^{K}$
and the $d$-th row of their weight matrix $\boldsymbol{W}$. The
1-head GAT was proposed for node classification problems. When it
is applied to graph learning tasks, requiring the $\boldsymbol{C}^{(l,d)}$
to be a rank-1 matrix severely limits the model capability and leads
to inferior performance.

\begin{figure*}
{\small{}}%
\begin{tabular}{c|cccccc}
\multicolumn{1}{c}{\textbf{\small{}Model Type }} & \textbf{\small{}MUTAG } & \textbf{\small{}PTC } & \textbf{\small{}PROTEINS } & \textbf{\small{}NCI1 } & \textbf{\small{}IMDB (B) } & \textbf{\small{}REDDIT (B) }\tabularnewline
\hline 
{\small{}Ego-CNN} & \textbf{\small{}93.1}{\small{} (6s) } & \textbf{\small{}63.8}{\small{} (20s) } & \textbf{\small{}73.8}{\small{} (244s) } & \textbf{\small{}80.7}{\small{} (854s) } & \textbf{\small{}72.3}{\small{} } & {\small{}87.8 }\tabularnewline
\hline 
{\small{}Patch-San + Std. Convolution Layers} & {\small{}89.4 } & {\small{}61.5 } & {\small{}70.7 } & {\small{}71.0 } & {\small{}67.1 } & {\small{}81.0}\tabularnewline
\hline 
{\small{}Ego-CNN with Scale-Free Regularizer } & {\small{}84.5 (13s) } & {\small{}59.5 (22s) } & {\small{}73.2 (281s) } & {\small{}77.8 (368s) } & {\small{}71.5 } & \textbf{\small{}88.4}{\small{} }\tabularnewline
\hline 
\end{tabular}{\small\par}

\caption{\label{fig:More-experiments}More experiments.}
\end{figure*}

\section{Detailed Visualization Steps}

In this section, we detail how the critical structure is visualized
using an Ego-CNN. Like standard CNN, each neuron in Ego-CNN represents
the matched result for a specific pattern (subgraph) with size upper-bounded
by $K^{L}$ nodes. To visualize the detected critical structure, the
first step is to to find the most important neighborhoods at the deepest
Ego-Convolution layer. The importance score $\gamma^{(n)}$ for each
node $n$ can be calculated using the Attention layer described in
Section 3.2 of the main paper. Next, we compute the importance-adjusted
node embedding $\boldsymbol{\hat{H}}_{n,:}^{(l)}=\gamma^{(n)}\hat{\boldsymbol{H}}_{n,:}^{(l)}\in\mathbb{R}^{D}$
before applying the Transposed Convolution  layer-by-layer (from
the deepest to the shallowest) to plot the critical subgraphs. At
each layer, we (1) follow the Transposed Convolution to re-construct
each node\textquoteright s importance-adjusted neighboring embeddings
$\hat{\boldsymbol{E}}^{(n,l)}=\sum_{d}\hat{\boldsymbol{H}}_{n,d}^{(l)}\boldsymbol{W}^{(l,d)}\in\mathbb{R}^{(K+1)\times D}$,
and then (2) reconstruct $\boldsymbol{\hat{H}}_{n,:}^{(l-1)}$ by
``undoing'' $\boldsymbol{E}^{(n,l)}=\left[\boldsymbol{H}_{n,:}^{(l-1)},\boldsymbol{H}_{Nbr(n,1),:}^{(l-1)},\cdots,\boldsymbol{H}_{Nbr(n,K),:}^{(l-1)}\right]^{\top}$
in Eq. (3) in the main paper. We do so by letting $\boldsymbol{\hat{H}}_{n,:}^{(l-1)}=\sum_{i:\text{node }i\text{ is }n\text{\textquoteright s }k\text{-th neightbor, }k\leq K}\hat{\boldsymbol{E}}_{k+1,:}^{(i,l)}$.
Repeating the above steps, we end up reconstructing an importance-adjusted
adjacency matrix $\hat{\boldsymbol{H}}^{(0)}\in\mathbb{R}^{k\times k}$
that ``undoes'' Patchy-San.  Our Figures 5 and 6 in the main paper
plot the graphs using edge widths proportional the values in $\hat{\boldsymbol{H}}^{(0)}$.

\section{More Experiments}

In this section, we conduct more experiments to further investigate
the performance of Ego-CNNs. First, we compare the Ego-CNN with a
naive extension of Patch-San using standard CNN layers. The architecture
(6 layers) and numbers of parameters of the two models are roughly
the same. The results are shown in Figure \ref{fig:More-experiments}.
As we can see, the Ego-CNN with Ego-Convolutions outperforms standard
CNN convolutions on graph classification problems. 

Next, we compare the Ego-CNNs with and without the scale-free regularizer
mentioned in Section 4.3 of the main paper. As we can see in Figure
\ref{fig:More-experiments}, the scale-free regularizer does not boost
performance on bioinformatic datasets (MUTAG/PTC/PROTEINS/NCI1) because
the graphs have no scale-free property. This motivates a test like
the one shown in Figure 5 in the main paper\textemdash one should
verify if the target graphs indeed have scale-free properties before
applying the scale-free regularizer.

\section{Relation to Kronecker Graphs}

In this section, we brief show how Kronecker graph \cite{leskovec2010kronecker}
is a special case of Ego-CNN with scale-free regularizer. Suppose
only 1 pattern $A^{(0)}\in\mathbb{{R}}^{K\times K}$ is captured in
the node embedding layer and there is only 1 filter $W\in\mathbb{{R}}^{(K+1)\times1}$in
the weight-tying Ego-Convolution layers. Then, the pattern captured
in the $l$-th weight-tying Ego-Convolution filter is $A^{(l)}=f(A^{(l-1)}\bigotimes W)$,
where $f$ is a function to convert the list of adjacency matrices
of length $K+1$into 1 unified adjacency matrix, where elements that
belong to the same physical nodes are merged.

\end{document}